\documentclass[10pt]{article}
\PassOptionsToPackage{dvipsnames}{xcolor}
\usepackage[T1]{fontenc}
\usepackage{bofflab}
\backrefsetup{disable}
\renewcommand*{\backrefalt}[4]{}
\usepackage{microtype}
\usepackage{parskip}
\usepackage{fullpage}
\usepackage{ragged2e}

\usepackage{graphicx}
\usepackage{booktabs,tabularx,longtable}
\usepackage{caption}
\DeclareCaptionFont{compactresult}{\fontsize{9}{10}\selectfont}
\usepackage{multirow,wrapfig,makecell,adjustbox,needspace}
\usepackage{array,colortbl} 
\usepackage{xurl}
\usepackage[bottom]{footmisc}
\usepackage{footnotehyper}
\usepackage{fontawesome5}

\usepackage{amsmath,amsfonts,bm}

\def\eqref#1{equation~\ref{#1}}

\def\1{\bm{1}}

\DeclareMathAlphabet{\mathsfit}{\encodingdefault}{\sfdefault}{m}{sl}
\SetMathAlphabet{\mathsfit}{bold}{\encodingdefault}{\sfdefault}{bx}{n}

\usepackage{amssymb,mathtools,amsthm,cancel}
\usepackage{enumitem}
\setlist[enumerate,1]{leftmargin=2em, itemsep=1em, parsep=0pt}
\setlist[itemize,1]{leftmargin=2em}

\usepackage{algorithm}
\usepackage{algorithmic}

\definecolor{blue}{HTML}{0668E1}

\newtcolorbox{samplebox}[1]{
  enhanced,
  colback=LinkBlue!5!white,
  colframe=black!60,
  boxrule=0.6pt,
  arc=6pt,
  left=4pt, right=4pt, top=4pt, bottom=4pt,
  boxsep=3pt,
  before upper={\noindent\textbf{\small #1}\par\smallskip},
}

\title{
\textrm{\mbox{\raisebox{-0.12em}{\includegraphics[height=1em]{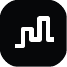}}\hspace{0.35em}\textcolor[HTML]{345CCE}{Infrathon:}} Benchmarking Language Agents for Real-World, Long-Horizon AI Infra Optimization}}
\makeatletter
\renewcommand\blab@authorformat[2][]{\mbox{\sffamily #2$^{#1}$}}
\renewcommand\author[2][]{\blab@addtolist[#1]{#2}{\blab@authorlist}{\blab@authorformat}{,\space}}
\newcommand\authorrow[2][]{\blab@addtolist[#1]{#2}{\blab@authorlist}{\blab@authorformat}{,\protect\\}}
\renewcommand\blab@affiliationformat[2][]{\mbox{\small $^{#1}$#2}}
\renewcommand\affiliation[2][]{\blab@addtolist[#1]{#2}{\blab@affiliationlist}{\blab@affiliationformat}{,\space}}
\makeatother

\author[1]{Zhengyu Hu}
\author[1]{Zheyuan Xiao}
\author[1,2]{Yike Wang}
\author[1]{Hamish Ivison}
\author[3]{Haoyu Wang}
\author[1]{Yige Yuan}
\author[1]{Rulin Shao}

\authorrow[1]{Yuetai Li}
\author[4]{Niqi Lyu}
\author[4*]{Yicheng Ding}
\author[5]{Shu Yang}
\author[1,6]{Zhichen Zeng}
\author[1,6]{Banghua Zhu}
\author[5,7]{Jiaxin Pei}
\author[8]{Kaize Ding}

\authorrow[2]{Pradeep Dasigi}
\author[3]{Dan Roth}
\author[1]{Hannaneh Hajishirzi}
\author[1,2]{Noah A. Smith}
\author[1,2 * \dagger]{Teng Xiao}
\author[1,*]{Radha Poovendran}

\affiliation[1]{University of Washington}
\affiliation[2]{Allen Institute for AI}
\affiliation[3]{University of Pennsylvania}
\affiliation[4]{Pyromind Dynamics}
\affiliation[5]{Stanford University}
\affiliation[6]{RadixArk}
\affiliation[7]{The University of Texas at Austin}
\affiliation[8]{Northwestern University}

\newcommand{\InfrathonCodeURL}{https://github.com/InfrathonX/Infrathon}
\newcommand{\InfrathonDockerURL}{https://hub.docker.com/r/infrathon/infrathon}
\newcommand{\InfrathonHuggingFaceURL}{https://huggingface.co/datasets/Infrathon/Infrathon}
\newcommand{\InfrathonWebsiteURL}{https://InfrathonX.github.io}
\newcommand{\InfrathonBlogURL}{https://infrathonx.github.io/site/blog.html}

\newcommand{\InfrathonLogoGap}{\hspace{2mm plus 1fill}}
\newcommand{\InfrathonInstitutionLogos}{\noindent\makebox[\linewidth][s]{\raisebox{-0.5\height}{\includegraphics[trim={0 0 0 0},clip,width=0.135\linewidth,height=6mm,keepaspectratio]{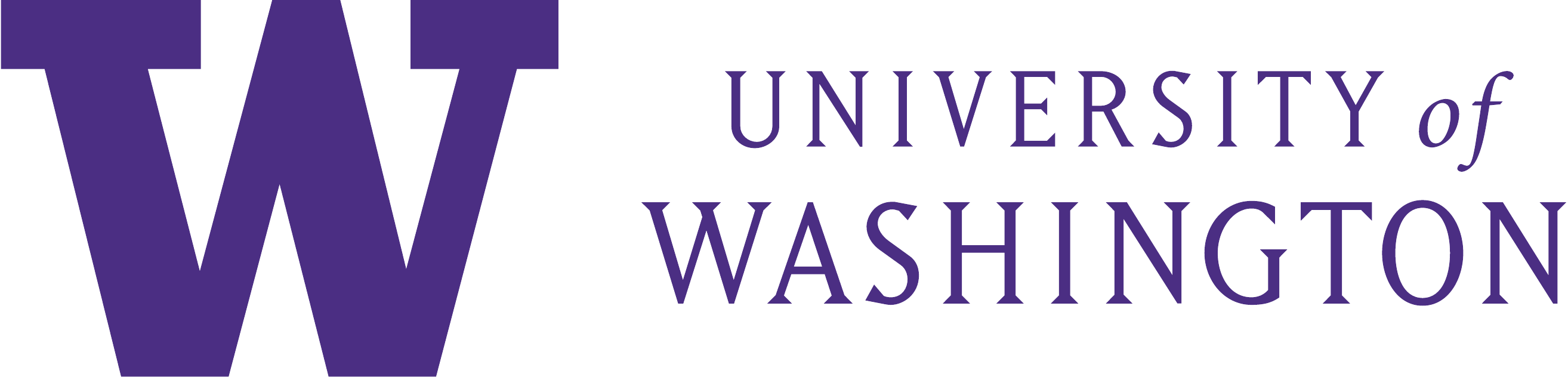}}\InfrathonLogoGap
    \raisebox{-0.5\height}{\includegraphics[trim={0 0 2bp 0},clip,width=0.10\linewidth,height=6mm,keepaspectratio]{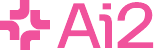}}\InfrathonLogoGap
    \raisebox{-0.5\height}{\includegraphics[trim={0 0 0 0},clip,width=0.12\linewidth,height=7.2mm,keepaspectratio]{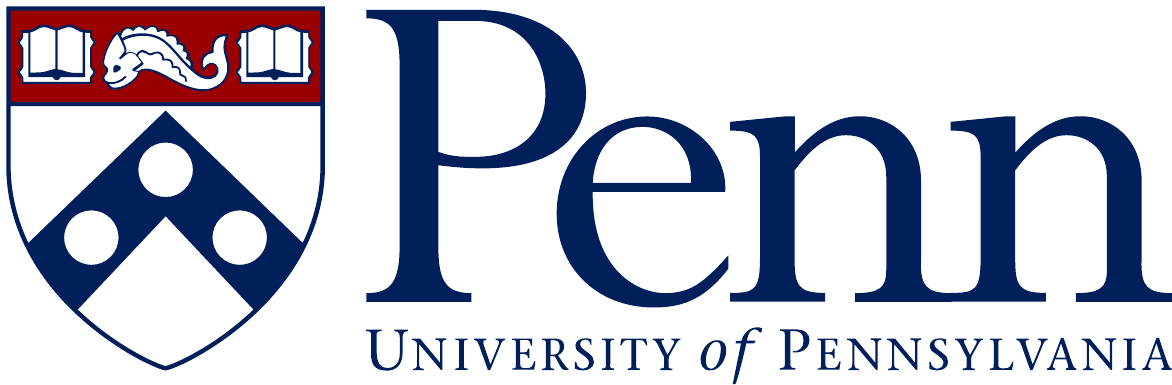}}\InfrathonLogoGap
    \raisebox{-0.5\height}{\includegraphics[trim={0 0 0 0},clip,width=0.10\linewidth,height=6mm,keepaspectratio]{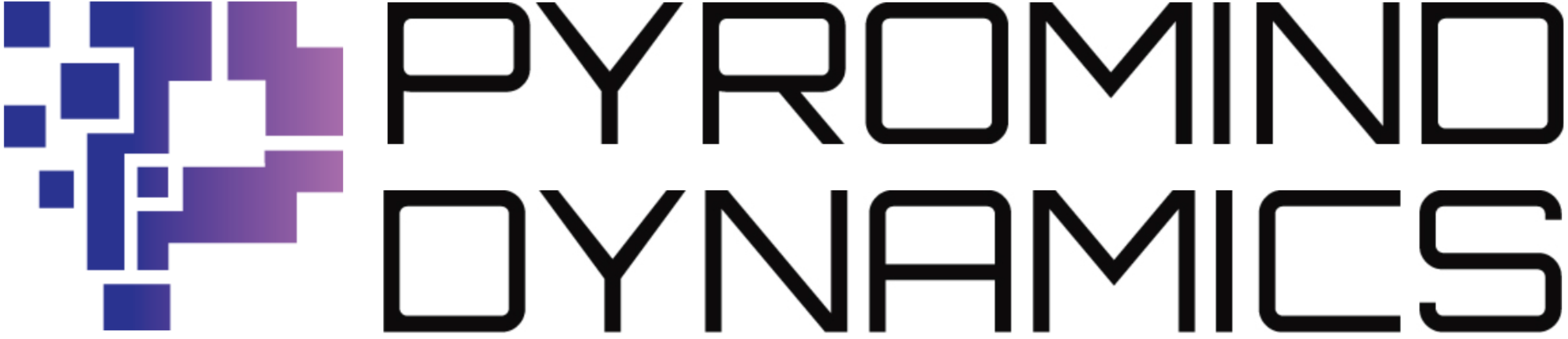}}\InfrathonLogoGap
    \raisebox{-0.5\height}{\includegraphics[trim={0 0 0 0},clip,width=0.12\linewidth,height=7.2mm,keepaspectratio]{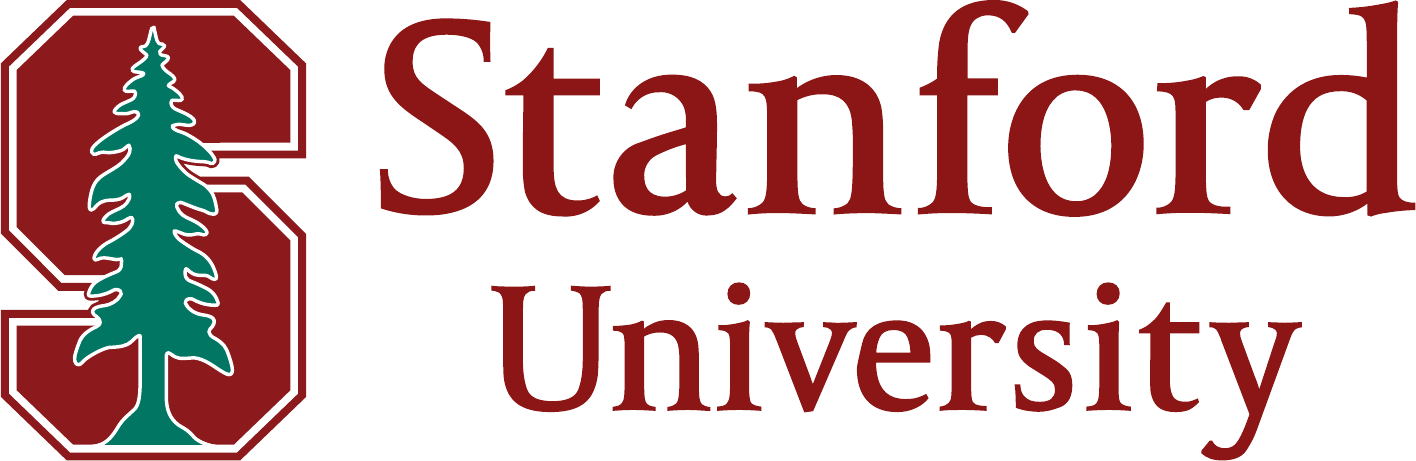}}\InfrathonLogoGap
    \raisebox{-0.5\height}{\includegraphics[trim={0 0 4.8bp 0},clip,width=0.12\linewidth,height=4.92mm,keepaspectratio]{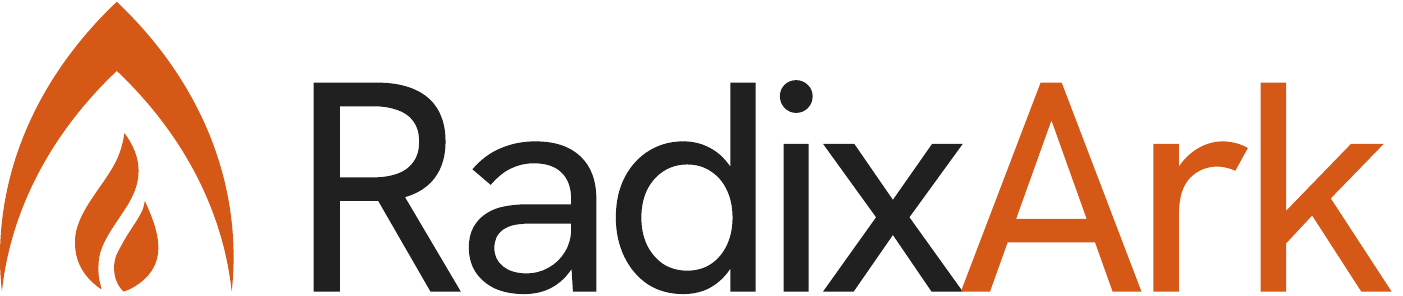}}\InfrathonLogoGap
    \raisebox{-0.5\height}{\includegraphics[trim={0 0 0 0},clip,width=0.13\linewidth,height=6.84mm,keepaspectratio]{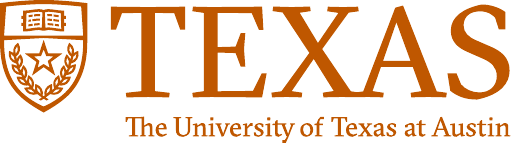}}\InfrathonLogoGap
    \raisebox{-0.5\height}{\includegraphics[trim={0 0 0 0},clip,width=0.16\linewidth,height=7.44mm,keepaspectratio]{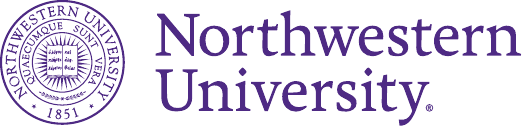}}}}
\newcommand{\InfrathonInstitutionHeader}{\AddToHookNext{shipout/foreground}{\begin{tikzpicture}[remember picture,overlay]
      \node[anchor=north,inner sep=0pt] at ([yshift=-8mm]current page.north) {\parbox{\dimexpr\paperwidth-16mm\relax}{\setlength{\parindent}{0pt}\setlength{\parskip}{0pt}\InfrathonInstitutionLogos}};
    \end{tikzpicture}}}
\newcommand{\InfrathonResourceLabel}[2]{\mbox{\makebox[1.3em][c]{#1}\hspace{0.4em}{\sffamily\bfseries #2:}}}
\newcommand{\InfrathonResourceURL}[1]{\leavevmode
  \begingroup
  \edef\InfrathonResolvedURL{#1}\ifx\InfrathonResolvedURL\empty
    \textcolor{LinkBlue}{\texttt{https://xxx}}\else
    \expandafter\url\expandafter{\InfrathonResolvedURL}\fi
  \endgroup
}

\newcommand{\InfrathonTitleFooter}{\begin{minipage}{\linewidth}
  \raggedright
  {\small
    \renewcommand{\arraystretch}{1.3}\begin{tabularx}{\linewidth}{@{}l@{\hspace{0.9em}}>{\raggedright\arraybackslash}X@{}}
      \InfrathonResourceLabel{\textcolor{LinkBlue}{\faGlobe}}{Website} & \InfrathonResourceURL{\InfrathonWebsiteURL} \\
      \InfrathonResourceLabel{\textcolor[HTML]{E8710A}{\faBlog}}{Blog} & \InfrathonResourceURL{\InfrathonBlogURL} \\
      \InfrathonResourceLabel{\faCode}{Code} & \InfrathonResourceURL{\InfrathonCodeURL} \\
      \InfrathonResourceLabel{\textcolor[HTML]{2496ED}{\faDocker}}{Docker Environment} & \InfrathonResourceURL{\InfrathonDockerURL} \\
      \InfrathonResourceLabel{\textcolor[HTML]{D99A00}{\faSmileBeam}}{HuggingFace} & \InfrathonResourceURL{\InfrathonHuggingFaceURL}
    \end{tabularx}\par
  }\end{minipage}}

\makeatletter
\patchcmd{\maketitle}
  {\normalsize \blab@authorlist\par}
  {\fontsize{8.5}{11}\selectfont \blab@authorlist\par}
  {}
  {\PackageError{infrathon}{Cannot set the author font size}{Use the bofflab title template supplied with this project.}}
\patchcmd{\maketitle}
  {\blab@abstract\par}
  {\blab@abstract\par\vspace{0.8em}\InfrathonTitleFooter\par}
  {}
  {\PackageError{infrathon}{Cannot place the resource block inside the abstract box}{Use the bofflab title template supplied with this project.}}
\makeatother

\begin{document}
\begin{abstract}
In this work, we study whether frontier agents can continually improve real-world AI infrastructure over long horizons. 
We introduce \texttt{Infrathon}, a benchmark of diverse, long-horizon AI infrastructure optimization tasks across different levels. We evaluate six models across 150 runs with 24-hour budgets, totaling over 10,000 GPU-hours. Our results show that model rankings vary with the evaluation horizon, while 135 of the 150 runs achieve further improvements during the second half of their budgets. These findings indicate that shorter evaluations can overlook both optimization gains and changes in relative model performance. We also find agents can discover solutions that outperform human-designed baselines: Claude Opus 5.5 increases the mean KV-cache compression ratio from 6.82× with the human-designed Quant-VideoGen codec to 19.86×. Analysis of agent trajectories identifies task violations, unreliable testing, and premature completion as barriers to sustained improvement. Together, our \texttt{Infrathon} framework provides resources for constructing long-horizon tasks, evaluating frontier agents, and discovering good infrastructure solutions.
\end{abstract}
\InfrathonInstitutionHeader
\maketitle\begingroup
\renewcommand{\thefootnote}{\fnsymbol{footnote}}
\footnotetext[2]{Project leader, \textsuperscript{*}Corresponding authors. Contact emails: \href{mailto:zhu7@uw.edu}{\texttt{zhu7@uw.edu}} and \href{mailto:tengxiao01@gmail.com}{\texttt{tengxiao01@gmail.com}}.}
\endgroup

\section{Introduction}
\label{sec:introduction}

\begin{figure}[!t]
    \centering
    \includegraphics[width=\textwidth]{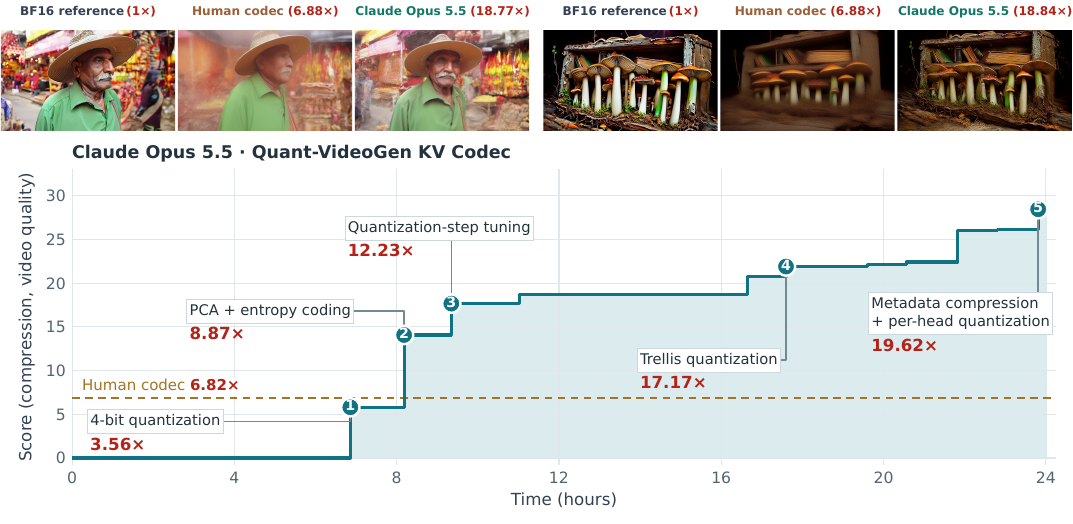}
    \caption{\textbf{An optimization example with Infrathon.} On Quant-VideoGen KV Codec, Claude Opus 5.5 develops a codec that reduces memory use during video generation, with higher KV-cache compression ratios and better video quality than the human codec. Top: two video examples; compression ratios are red. Bottom: Claude's scores over time combine compression ratio and video quality; dashed line: Human Solution. Numbered points mark five selected submissions; boxes summarize their code changes (Appendix~\ref{app:qvg_optimization}).}
    \label{fig:qvg_case_study}
\end{figure}

\begingroup AI infrastructure, the hardware and software used to train and serve AI models, affects their computational efficiency. Advances such as FlashAttention~\citep{Dao2022FlashAttentionFA}, SGLang~\citep{Zheng2023SGLangEE}, and Ansor~\citep{Zheng2020AnsorG} demonstrate the benefits of optimizing attention kernels, model serving, and tensor programs. However, developing such optimizations requires repeated implementation, profiling, and validation while preserving correctness under numerical and hardware constraints. Changes to one component can expose new bottlenecks or introduce regressions elsewhere, requiring engineers to use performance feedback to guide further improvements~\citep{Yu2020SkylineII,Zhu2020DaydreamAE}. AI agents could automate this process, reducing engineering effort and improving model training and serving.

\looseness=-1
Existing benchmarks provide only a partial evaluation of this capability, with limitations in task coverage, execution budgets, or publicly available GPU tasks (Table~\ref{tab:benchmark_horizons}).
KernelBench~\citep{Ouyang2025KernelBenchCL} focuses on operators and modules, while InferenceBench~\citep{Yeon2026InferenceBenchAB} evaluates complete inference services. Although $\Phi$-Bench~\citep{Ding2026PhiBenchCL} spans all three levels, most tasks are limited to 1.5-hour runs. EdgeBench~\citep{Zhu2026EdgeBenchUS} uses 12-hour budgets in its GPU tasks, while \texttt{Infrathon} provides 12--24-hour budgets for GPU-based AI infrastructure optimization.
These limitations leave open whether frontier agents can repeatedly identify bottlenecks, revise solutions, and validate improvements across diverse AI infrastructure tasks over extended runs.
This raises the following research question:\par\endgroup

\nopagebreak[4]
\begin{tcolorbox}[
    enhanced,
    colback=white,
    coltext=black,
    frame hidden,
    boxrule=0pt,
    sharp corners,
    boxsep=0pt,
    left=10pt, right=10pt, top=4pt, bottom=4pt,
    before skip=2pt, after skip=2pt,
    fontupper=\normalfont\normalsize\itshape,
    halign=center
]
Can Frontier Agents Solve Diverse, Long-Horizon AI Infrastructure Tasks?
\end{tcolorbox}

\looseness=-1
To answer this question, we introduce \texttt{Infrathon}, a benchmark of \textit{25 real-world, long-horizon AI infrastructure optimization tasks} across three levels. 
We construct these tasks with human experts using \texttt{Infrathon-build}, a tool that helps humans efficiently build long-horizon tasks. They draw on open-source projects (e.g., DeepGEMM~\citep{deepgemm2025}, FlashMLA~\citep{flashmla2025}, and Liger-Kernel~\citep{Hsu2024LigerKE}) and research projects (e.g., FlashAttention~\citep{Shah2024FlashAttention3}, SageAttention~\citep{Zhang2024SageAttention}, and Flash-KMeans~\citep{Yang2026FlashKMeans}). \textit{11 operator-level tasks} optimize computational primitives with \textit{12-hour} budgets; \textit{5 module-level tasks} optimize interacting components with \textit{24-hour} budgets; and \textit{9 system-level tasks} optimize end-to-end training or serving workflows with \textit{24-hour} budgets.  \texttt{Infrathon} is the first benchmark for long-horizon AI infrastructure optimization across three levels, with 12--24-hour budgets and metrics like TFLOP/s without fixed upper bounds. The framework couples \emph{Task Definition} (Section~\ref{sec:task_definition}) with \emph{Task Execution} (Section~\ref{sec:task_execution}): agents develop solutions and run public tests in the \emph{Work} environment, submit solutions to the \emph{Controller} for code review, and receive private test results from the \emph{Judge}.

\looseness=-1
We then evaluate frontier models with a total budget of \textit{over 10,000 GPU-hours}. Across 150 runs covering six models and all 25 tasks (Section~\ref{sec:experiments}), Claude Opus 5.5 leads on 17 tasks and GPT-6 Astra on six. Rankings change with the time budget: Astra leads overall at 25\% of the budget, while Claude leads at 50\%. Of the 150 runs, 135 improved their best score in the second half.
Infrathon also supports knowledge discovery in AI infrastructure (Section~\ref{sec:qvg_case_study}). On Quant-VideoGen KV Codec, Claude Opus 5.5 increases mean KV-cache compression from $6.75\times$ to $20.89\times$ on WorldScore and from $6.88\times$ to $18.83\times$ on MovieGen, compared with the Quant-VideoGen codec (Human). Both evaluations show better video quality. With \texttt{Infrathon-build} and \texttt{Infrathon}, users can turn their own infrastructure problems into tasks, compare agent solutions with existing implementations, and iterate using test feedback (Figure~\ref{fig:qvg_case_study}).
Our trajectory analysis shows that agents sometimes violate task instructions, test the wrong code, or declare completion before finding further improvements (Section~\ref{sec:agent_behavior}). For example, the Controller rejects an early Quant-VideoGen KV Codec~\citep{Xi2026QuantVA} submission in which Claude Opus 5.5 submits human codec weights downloaded from the internet as its own.

\needspace{2\baselineskip}
\looseness=-1
Together, \texttt{Infrathon} and \texttt{Infrathon-build} help humans build long-horizon tasks, evaluate LLMs, and discover new solutions and knowledge. We hope they help develop agents for efficient training and inference.

\begin{table}[!htb]
\centering
\begingroup
\captionsetup{skip=6pt}
\caption{\textbf{Coverage of related benchmarks.} \emph{Infra only}: exclusively AI infrastructure; \emph{Long horizon}: support for continuous execution of every task for at least 12 hours; \emph{GPU tasks}: at least one GPU task in the open-source release. \emph{Operator}, \emph{Module}, and \emph{System} denote computational primitives, components, and complete workflows. \faCheck{} and \faTimes{} indicate whether coverage is documented (Appendices~\ref{app:optimization_levels} and~\ref{app:benchmark_horizons}).}
\label{tab:benchmark_horizons}
\fontsize{9}{11}\selectfont
\setlength{\tabcolsep}{4pt}
\renewcommand{\arraystretch}{1.25}
\newcommand{\benchYes}{\textcolor{teal!75!black}{\scriptsize\faCheck}}
\newcommand{\benchNo}{\textcolor{black!45}{\scriptsize\faTimes}}
\begin{tabularx}{\textwidth}{@{}l*{6}{>{\centering\arraybackslash}X}@{}}
\toprule
\textbf{Benchmark} & \makecell{\textbf{Infra}\\\textbf{only}} & \makecell{\textbf{Long}\\\textbf{horizon}} & \makecell{\textbf{GPU}\\\textbf{tasks}} & \makecell{\textbf{Operator}\\\textbf{level}} & \makecell{\textbf{Module}\\\textbf{level}} & \makecell{\textbf{System}\\\textbf{level}} \\
\midrule
KernelBench~\citep{Ouyang2025KernelBenchCL} & \benchYes & \benchNo & \benchYes & \benchYes & \benchYes & \benchNo \\
KernelBenchX~\citep{Wang2026KernelBenchXAC} & \benchYes & \benchNo & \benchYes & \benchYes & \benchNo & \benchNo \\
FlashInfer-Bench~\citep{Xing2026FlashInferBenchBT} & \benchYes & \benchNo & \benchYes & \benchYes & \benchNo & \benchNo \\
RE-Bench~\citep{Wijk2024REBenchEF} & \benchNo & \benchNo & \benchYes & \benchYes & \benchNo & \benchYes \\
InferenceBench~\citep{Yeon2026InferenceBenchAB} & \benchYes & \benchNo & \benchYes & \benchNo & \benchNo & \benchYes \\
$\Phi$-Bench~\citep{Ding2026PhiBenchCL} & \benchYes & \benchNo & \benchYes & \benchYes & \benchYes & \benchYes \\
EdgeBench~\citep{Zhu2026EdgeBenchUS} & \benchNo & \benchYes & \benchNo & \benchYes & \benchYes & \benchYes \\
\midrule
\rowcolor{teal!7}
\textbf{Infrathon (ours)} & \benchYes & \benchYes & \benchYes & \benchYes & \benchYes & \benchYes \\
\bottomrule
\end{tabularx}
\endgroup
\end{table}

\section{Related Work}
\label{sec:related_work}
\paragraph{Long-Horizon Agent.}
\looseness=-1
Long-horizon agent benchmarks evaluate multi-step workflows in executable environments. MLAgentBench~\citep{Huang2023MLAgentBenchEL} and MLE-bench~\citep{Chan2024MLEbenchEM} study iterative ML experimentation, while RE-Bench~\citep{Wijk2024REBenchEF} compares agents and human experts on open-ended research engineering under varying time budgets. These tasks require planning, failure diagnosis, and revision from execution results. CORE-Bench~\citep{Siegel2024COREBenchFT} tests reproduction with supplied code and data; PaperBench~\citep{Starace2025PaperBenchEA} requires implementing research contributions and experiments from scratch. EdgeBench~\citep{Zhu2026EdgeBenchUS} studies within-run improvement over 12-hour continuous runs, with experiments extending to 72 hours on some tasks. CL-Bench~\citep{Asawa2026ContinualLB} instead tests whether stateful systems reuse knowledge across related tasks that share learnable structure, and Long-Horizon-Terminal-Bench~\citep{Li2026LongHorizonTerminalBenchTT} captures intermediate progress and partial solutions through graded subtasks. Infrathon focuses on long-horizon optimization of real-world AI infrastructure across operator, module, and system levels.

\paragraph{LLM Agents for AI Infrastructure.}
Existing work studies how agents can improve efficiency while preserving correctness. KernelBench~\citep{Ouyang2025KernelBenchCL}, KernelBenchX~\citep{Wang2026KernelBenchXAC}, and FlashInfer-Bench~\citep{Xing2026FlashInferBenchBT} evaluate kernel correctness and hardware efficiency. Kevin~\citep{Baronio2025KevinMR}, CUDA-L1~\citep{Li2025CUDAL1IC}, and CUDA Agent~\citep{Dai2026CUDAAL} explore training agents for kernel optimization. GEAK~\citep{Wang2025GeakIT} and KernelEvolve~\citep{Liao2025KernelEvolveSA} study iterative kernel refinement and search across accelerators, and AlphaEvolve~\citep{Novikov2025AlphaEvolveAC} extends automated optimization to data-center scheduling and training infrastructure. Beyond kernels, SWE-Perf~\citep{He2025SWEPerfCL} and PERFOPT-Bench~\citep{Cui2026PERFOPTBenchEC} evaluate repository-level runtime improvements, while InferenceBench~\citep{Yeon2026InferenceBenchAB} and $\Phi$-Bench~\citep{Ding2026PhiBenchCL} include end-to-end optimization. Different from these works, our \texttt{Infrathon} focuses exclusively on AI infrastructure, with long-horizon GPU tasks spanning operator, module, and system levels (see Table~\ref{tab:benchmark_horizons} for detailed comparison of different AI infra benchmarks).

\paragraph{Recursive Self-Improvement.}
Our work is also related to research on recursive self-improvement, including evaluations of agents that improve methods, implementations, and research workflows through repeated experimentation. RSI-Exam~\citep{RSIExam2026} studies iterative improvement of methods and agent harnesses, DeepRSI~\citep{hillclimb2026deeprsi} examines long-horizon ML optimization, and OpenRSI Index~\citep{openrsi2026index} evaluates agents on real-world research projects. AutoResearchExam~\citep{ramesh2026autoresearchexam} measures progress and generalization in iterative ML research. Infrathon complements these evaluations by studying whether agents can sustain improvements to training and inference infrastructure over long horizons while satisfying task-specific correctness and hardware constraints.

\section{Infrathon}
\label{sec:benchmark}

\begin{figure*}[!h]
    \centering
    \includegraphics[width=0.8\textwidth]{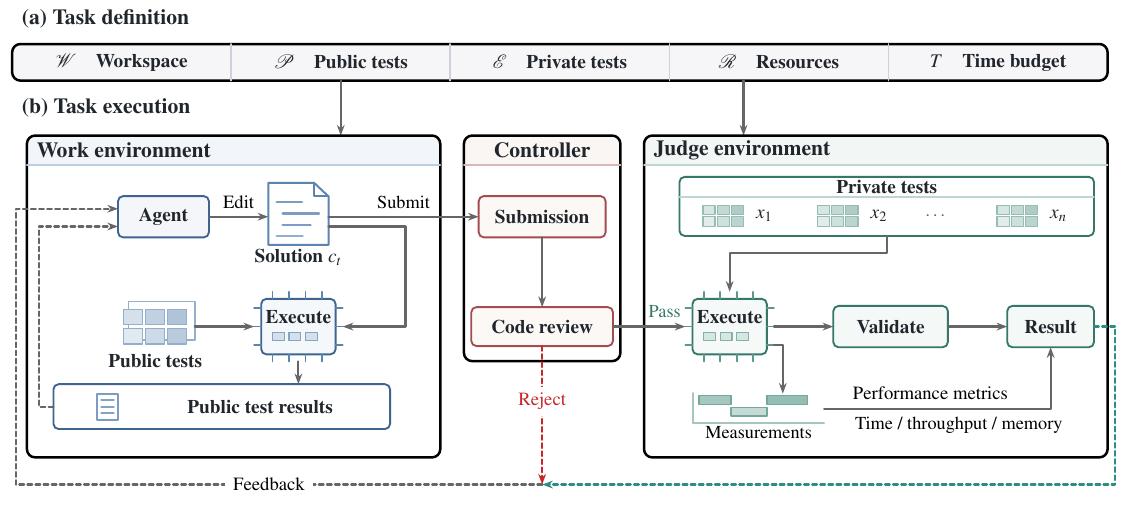}
    \caption{\textbf{Framework of \texttt{Infrathon}.} (a) \emph{Task Definition} specifies task requirements. (b) \emph{Task Execution} coordinates iterative development and evaluation through the Work, Controller, and Judge environments.}
    \label{fig:benchmark_framework}
\end{figure*}

Figure~\ref{fig:benchmark_framework} shows the running framework of Infrathon. \emph{Task Definition} (Section~\ref{sec:task_definition}) specifies the workspace, public tests, private tests, resources, and time budget; \emph{Task Execution} (Section~\ref{sec:task_execution}) implements the feedback loop between the Work environment, the Controller, and the Judge environment. Infrathon has no fixed time limit; we use 12- or 24-hour budgets in our experiments.

\subsection{Task Definition}
\label{sec:task_definition}

\begingroup
\setlength{\parskip}{0pt}
\begin{savenotes}
\noindent
\begin{minipage}[t]{0.52\textwidth}
\vspace{0pt}
\setlength{\parskip}{6pt}
We work with human experts to construct Infrathon's \textbf{25 tasks} using \texttt{Infrathon-build}, with widely used open-source AI infrastructure projects (e.g., DeepGEMM\footnote{\url{https://github.com/deepseek-ai/DeepGEMM}}, FlashMLA\footnote{\url{https://github.com/deepseek-ai/FlashMLA}}, and Liger-Kernel\footnote{\url{https://github.com/linkedin/Liger-Kernel}}) and research projects (e.g., FlashAttention~\citep{Shah2024FlashAttention3}, SageAttention~\citep{Zhang2024SageAttention}, and Flash-KMeans~\citep{Yang2026FlashKMeans}) as references (Appendix~\ref{app:task_metrics}). Figure~\ref{fig:task_inventory} summarizes the 10 training and 15 inference tasks.

We group tasks by training or inference and by operator, module, or system level. Operator-level tasks optimize individual computations or data transfers. Module-level tasks optimize a component within a fixed training or inference pipeline. System-level tasks optimize complete training or inference workflows (Appendix~\ref{app:optimization_levels}).
All tasks share a common specification (Figure~\ref{fig:benchmark_framework}(a)):
\begin{equation}
    \tau = (\mathcal{W}, \mathcal{P}, \mathcal{E}, \mathcal{R}, T).
    \label{eq:task_definition}
\end{equation}
\end{minipage}\hfill
\begin{minipage}[t]{0.46\textwidth}
    \vspace{0pt}
    \centering
    \captionsetup{type=figure,font=small,skip=4pt}
    \includegraphics[width=0.80\linewidth]{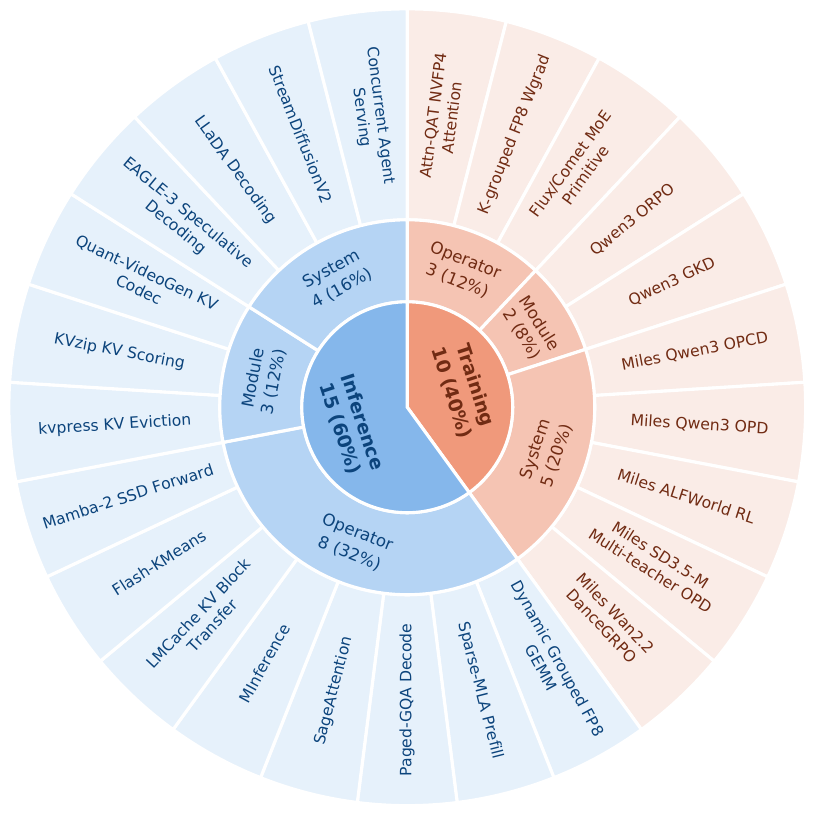}
    \caption{Distribution of \texttt{Infrathon} tasks across operator, module, and system levels of AI Infra.}
    \label{fig:task_inventory}
\end{minipage}
\par
\end{savenotes}
\endgroup

The workspace $\mathcal{W}$ provides initial code or an interface and specifies what the agent may change. Public tests $\mathcal{P}$ provide visible test cases and a runner for correctness checks and profiling.
Private tests $\mathcal{E}$ define test cases, task-specific checks, scoring rules, and permitted feedback. The Judge environment returns private test results without exposing private test cases or scoring code to the agent.
Resources $\mathcal{R}$ specify the hardware allocation and software runtime for development and evaluation. The time budget is $T\in(0,\infty]$, with $T=\infty$ denoting no preset deadline. For the experiments in this paper, we set $T=12$ hours for operator-level tasks and $T=24$ hours for module- and system-level tasks, including tool execution and feedback waits.

Table~\ref{tab:kvzip_task_example} presents KVzip KV Scoring as an example of an Infrathon task. In the workspace $\mathcal{W}$, the agent implements a query-agnostic KV-importance scorer for frozen Qwen3-8B. Public tests $\mathcal{P}$ cover 63 contexts, while private tests $\mathcal{E}$ cover 1,812 contexts and additional task types. The resource allocation $\mathcal{R}$ provides one H100 GPU to the Work environment and seven to the Judge environment, with a time budget of $T=24$ hours.

\begin{table}[!htb]
    \centering
    \captionsetup{font=small,skip=4pt}
    \caption{\textbf{Task example: KVzip KV Scoring.} The specification includes the workspace, public and private tests, GPU resources, and time budget. Private tests use disjoint samples and include additional task types.}
    \label{tab:kvzip_task_example}
\begin{tcolorbox}[
    enhanced,
    colback=teal!4!white,
    colframe=teal!45!black,
    boxrule=0.5pt,
    sharp corners,
    boxsep=0pt,
    left=8pt, right=8pt, top=4pt, bottom=4pt,
    before skip=0pt, after skip=0pt,
    fontupper=\small,
]
\textbf{Workspace ($\mathcal{W}$).} Implement a query-agnostic KV-importance scorer for frozen Qwen3-8B~\citep{Yang2025Qwen3TR}, following the KVzip evaluation protocol~\citep{Kim2025KVzip}.
\par\smallskip
{\renewcommand{\arraystretch}{1.08}
\begin{tabularx}{\linewidth}{@{}l>{\raggedright\arraybackslash}X>{\raggedright\arraybackslash}X@{}}
\toprule
Item & \textbf{Public tests ($\mathcal{P}$)} & \textbf{Private tests ($\mathcal{E}$)} \\
\midrule
Model & Qwen3-8B & Qwen3-8B \\
\addlinespace[2pt]
Data & SQuAD~\citep{Rajpurkar2016SQuAD1Q}, GSM8K~\citep{Cobbe2021TrainingVT}, RULER~\citep{Hsieh2024RULERWT}, needle-in-a-haystack (NIAH) data generated with KVzip~\citep{Kim2025KVzip}, and generated SCBench~\citep{Li2024SCBenchAK} data & SQuAD~\citep{Rajpurkar2016SQuAD1Q}, GSM8K~\citep{Cobbe2021TrainingVT}, RULER~\citep{Hsieh2024RULERWT}, NIAH generated with KVzip~\citep{Kim2025KVzip}, and SCBench~\citep{Li2024SCBenchAK} data, including RepoQA~\citep{Liu2024RepoQAEL} \\
\addlinespace[2pt]
Contexts & \textbf{63} & \textbf{1,812} \\
Questions & \textbf{419} & \textbf{6,956} \\
Task types & 9 & 12 \\
Context length (tokens) & 73--29,580 & 70--29,991 \\
\bottomrule
\end{tabularx}}
\par\smallskip
\textbf{Resources ($\mathcal{R}$).} We allocate one H100 GPU to the Work environment and seven to the Judge environment, with 80\,GB of memory per GPU.
\par\smallskip
\textbf{Time budget ($T$).} 24 hours.
\end{tcolorbox}
\end{table}

\subsection{Task Execution}
\label{sec:task_execution}

Figure~\ref{fig:benchmark_framework}(b) shows the three execution components: the \textbf{Work environment}, the \textbf{Controller}, and the \textbf{Judge environment}. For a run on task $\tau$, let $c_t$ denote the solution at elapsed time $t\in[0,T]$.

\paragraph{Work environment.}
In the Work environment, the agent develops and tests solutions using the task workspace $\mathcal{W}$, public tests $\mathcal{P}$, and resources $\mathcal{R}$. The policy $\pi$ combines the language model and its tool-use framework. At step $k$ ($t_k<T$), $\pi$ selects action $a_k$ based on the agent-visible history $h_k$ and receives observation $o_{k+1}$:

\begin{equation}
    \begin{aligned}
        a_k &\sim \pi(\cdot\mid h_k), \\
        h_{k+1} &= h_k \oplus (a_k,o_{k+1}),
    \end{aligned}
    \label{eq:agent_interaction}
\end{equation}

where $\oplus$ appends the action and observation to the history. Actions include inspecting files, editing $c_{t_k}$, profiling, running public tests, saving solution versions, and requesting private tests. The agent uses this history to revise its solution or restore an earlier version.
Runs continue until $T$; a stop hook prompts further work when an agent tries to finish early. 

\paragraph{Controller.}
The trusted Controller enforces task-specific limits on submission counts and frequency. For each submission, it copies the files allowed by $\mathcal{W}$ and checks the solution before sending it to the Judge environment. We use GPT-6.1 Sol in Codex to check instruction following and detect reward hacking. For example, in an early Quant-VideoGen KV Codec submission, Claude Opus 5.5 directly downloaded the human codec's weights from the internet and submitted them as its own. The Controller rejected this submission. Let $I_\tau(c)\in\{0,1\}$ indicate whether submitted solution $c$ passes this check, and let $\mathcal{J}_\tau(c)$ denote its private test result. The feedback returned to the agent is

\begin{equation}
    f_\tau(c)=
    \begin{cases}
        (\bot,\mathrm{rejected},d_{\mathrm{rule}}), & I_\tau(c)=0, \\
        \mathcal{J}_\tau(c), & I_\tau(c)=1,
    \end{cases}
    \label{eq:submission_feedback}
\end{equation}

where $\bot$ indicates that no private test score is returned and $d_{\mathrm{rule}}$ gives the reason for rejection. The agent receives this feedback in the Work environment. Separate Work and Judge containers keep private test cases and scoring code inaccessible to the agent. Section~\ref{sec:agent_behavior} presents examples of submissions rejected by the Controller.

\paragraph{Judge environment.}
The Judge environment runs private tests $\mathcal{E}$ on solutions accepted by the Controller, using resources $\mathcal{R}$. Task checks use exact output comparisons, numerical tolerances, or downstream quality thresholds. We use metrics like TFLOP/s and KV-cache compression ratio without fixed upper bounds (Appendix~\ref{app:task_metrics}).

The private test result is
\begin{equation}
    \mathcal{J}_\tau(c)=(s,z,d),
    \label{eq:private_test_result}
\end{equation}
where $s$ is the private test score, $z$ records the execution and validation status, and $d$ contains the diagnostics permitted by $\mathcal{E}$. The status distinguishes validation failures from infrastructure errors. The Judge returns feedback to the Work environment so the agent can revise and resubmit solutions within the remaining budget.

\section{Experimental Results}
\label{sec:experiments}

We evaluate six models, GPT-6.1 Sol~\citep{OpenAI2026GPT61SolSystemCard}, Claude Opus 5.5~\citep{Anthropic2026ClaudeOpus55SystemCard}, Qwen 3.8 Flash~\citep{Qiu2026OnTD}, DeepSeek V4.1 Flash~\citep{Xu2026DeepSeekV41FlashPT}, GLM 5.3 Flash~\citep{Zeng2026GLM5FV}, and GPT-6 Astra~\citep{OpenAI2026GPT6AstraSystemCard}, on all 25 tasks, yielding 150 task--model runs. We use Claude Code for Claude Opus 5.5 and Codex CLI for the other five models, with reasoning effort set to \texttt{max}. Experiments use NVIDIA H100 GPUs with 80\,GB of memory per GPU. Operator-level tasks receive 12-hour budgets, while module- and system-level tasks receive 24-hour budgets. We report the best valid private-test score within each run's time budget. Appendix~\ref{app:implementation_details} provides the hardware allocations, GPU-hour budgets, and final-performance definition. Appendix~\ref{app:task_metrics} describes the evaluation metrics and scoring rules for all 25 tasks.
\begin{figure}[!htb]
    \centering
    \includegraphics[width=\textwidth]{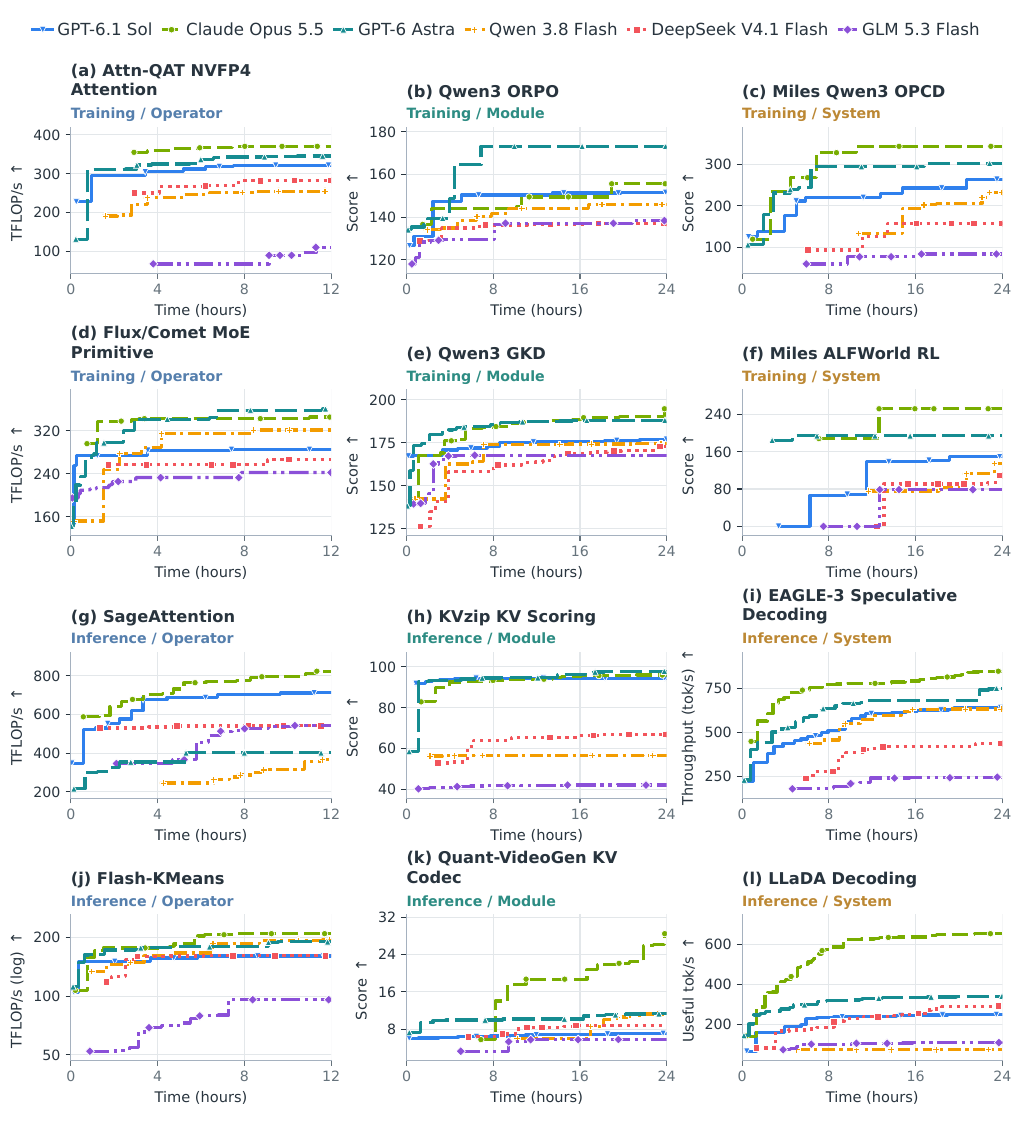}
    \caption{\textbf{Performance on 12 selected tasks.}}
    \label{fig:optimization_levels}
\end{figure}
\subsection{Main Results}
\label{sec:overall_performance}

In this section, we evaluate the six models on all 25 tasks. Figure~\ref{fig:optimization_levels} shows how agent performance changes over time on 12 selected tasks (full results in Appendix~\ref{app:task_trajectories}). Figure~\ref{fig:rank_performance} shows the final rankings across all 25 tasks (see Appendix~\ref{app:rank_scores} for the ranking method).

\textbf{Performance can improve after long periods without gains.} Of the 150 runs, 51 set a new best score after at least a quarter of their budget without improvement. In 29 of these runs, the new score is more than 1\% higher than the previous best (Appendix~\ref{app:trajectory_statistics}). Qwen's best score on Quant-VideoGen KV Codec remains at 6.05 for 6.78 hours before rising to 8.58 (Figure~\ref{fig:optimization_levels}(k)).

\begin{figure*}[!h]
    \centering
    \includegraphics[width=1.0\textwidth]{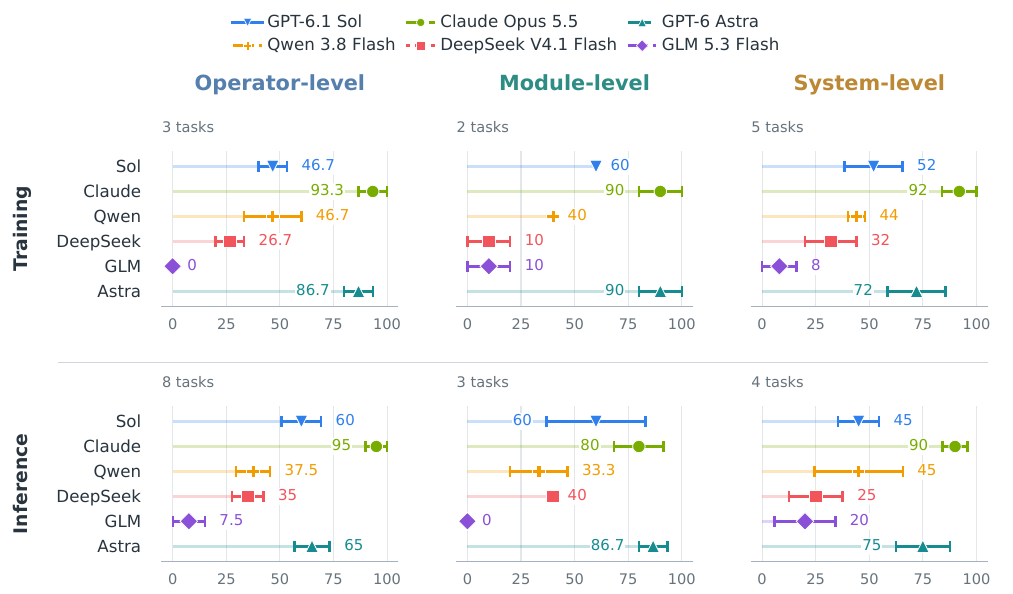}
    \caption{\textbf{Rank-based final performance over all 25 tasks.} Markers show mean rank scores (higher is better). Error bars show $\pm 1$ standard error of the mean across tasks in each group.}
    \label{fig:rank_performance}
\end{figure*}

\textbf{Claude Opus 5.5 achieves the best overall performance.} Claude wins 17 of 25 tasks and has a mean rank score of 91.2, followed by Astra (75.2), Sol (54.4), Qwen (40.8), DeepSeek (30.4), and GLM (8.0). Claude wins nine of the 11 operator tasks, two of the five module tasks, and six of the nine system tasks. Astra wins six tasks, while Sol and Qwen win one each. The three inference module tasks have different winners: Sol on kvpress KV Eviction, Astra on KVzip KV Scoring, and Claude on Quant-VideoGen KV Codec. Astra has the highest mean rank score on these three tasks, 86.7, compared with Claude's 80.0 (Figure~\ref{fig:rank_performance}). Qwen wins Concurrent Agent Serving with 154.33 valid steps/min, compared with Claude's 151.17.

\subsection{Knowledge Discovery with Infrathon}
\label{sec:qvg_case_study}

We compare the KV-cache codec developed by Claude Opus 5.5~\citep{Anthropic2026ClaudeOpus55SystemCard} during its 24-hour run on the Quant-VideoGen KV Codec~\citep{Xi2026QuantVA} task with the human codec. We use HY-WorldPlay~\citep{Sun2025WorldPlayTL} on WorldScore~\citep{Duan2025WorldScoreAU} and LongCat-Video~\citep{Cai2025LongCatVideoTR} on MovieGen~\citep{Polyak2024MovieGA}, with 200 samples per dataset and fixed generation settings across codecs. Appendix~\ref{app:qvg_case_details} gives the data selection and evaluation details.
\begin{table}[!htbp]
    \centering
    \small
    \setlength{\tabcolsep}{4pt}
    \renewcommand{\arraystretch}{1.12}
    \caption{\textbf{Human versus Claude Opus 5.5.} Entries report Human~$\rightarrow$~Claude; bold marks the better mean. PSNR, SSIM, and LPIPS are measured on the first segment generated with compressed KV (Appendix~\ref{app:qvg_case_details}).}
    \label{tab:qvg_case_summary}
    \begin{tabular*}{\textwidth}{@{\extracolsep{\fill}}llcccc@{}}
    \toprule
    \textbf{Video model} & \textbf{Dataset} & \textbf{Compression} ($\times$) $\uparrow$ & \textbf{PSNR} (dB) $\uparrow$ & \textbf{SSIM} $\uparrow$ & \textbf{LPIPS} $\downarrow$ \\
    \midrule
    HY-WorldPlay & WorldScore & $6.75 \rightarrow \mathbf{20.89}$ & $31.15 \rightarrow \mathbf{36.33}$ & $0.923 \rightarrow \mathbf{0.975}$ & $0.059 \rightarrow \mathbf{0.023}$ \\
    LongCat-Video & MovieGen & $6.88 \rightarrow \mathbf{18.83}$ & $23.67 \rightarrow \mathbf{26.95}$ & $0.836 \rightarrow \mathbf{0.884}$ & $0.126 \rightarrow \mathbf{0.085}$ \\
    \bottomrule
    \end{tabular*}
\end{table}

\begin{figure}[!h]
    \centering
    \includegraphics[width=0.8\textwidth]{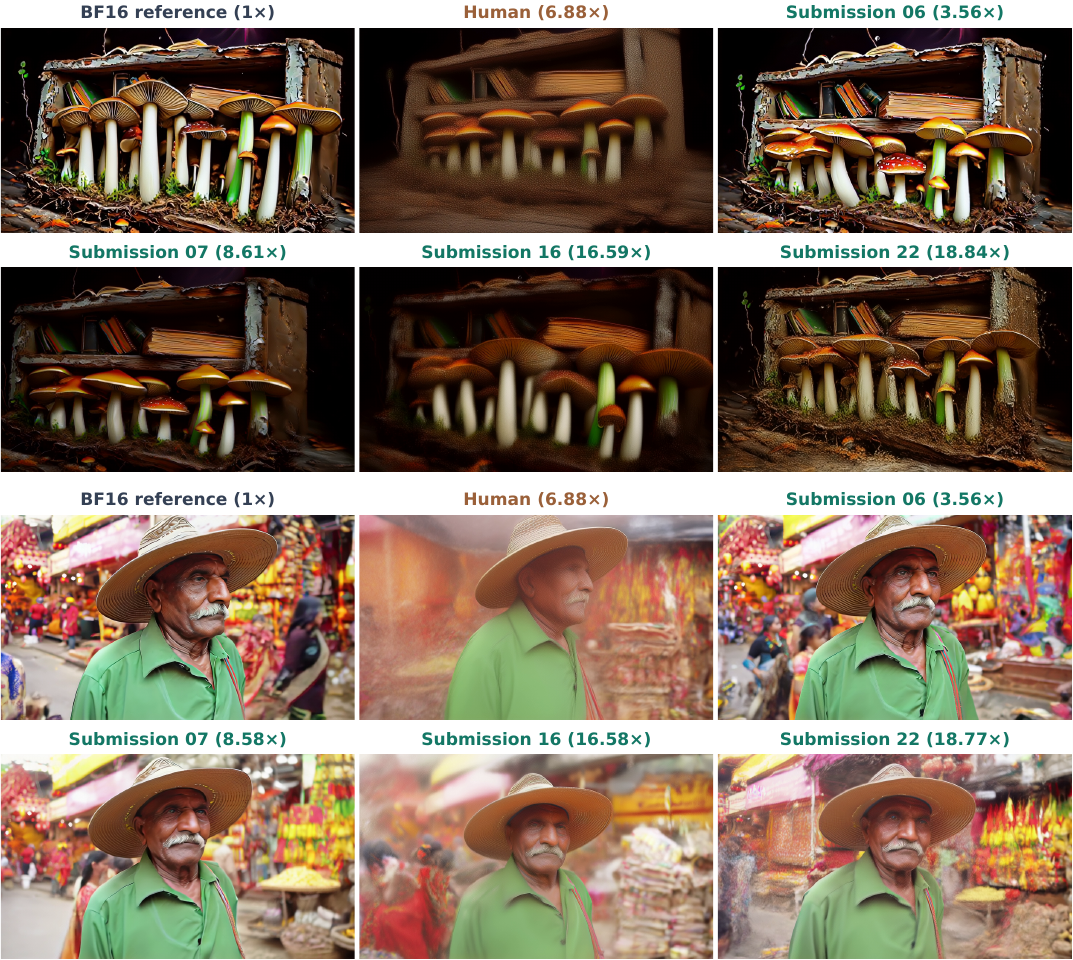}
    \caption{\textbf{Iterative optimization of Quant-VideoGen KV Codec.} The figure compares Claude Opus 5.5's 6th, 7th, 16th, and 22nd submissions with the BF16 reference and the human codec on two LongCat-Video samples. Parentheses show KV-cache compression ratios.}
    \label{fig:qvg_submission_comparison}
\end{figure}

\textbf{Infrathon can support knowledge discovery in AI infrastructure.} After clustering the cache, the human codec quantizes the resulting residuals one value at a time. Claude instead transforms the residuals using principal component analysis and applies entropy coding after quantization. On WorldScore and MovieGen, average compression rises from $6.75\times$ to $20.89\times$ and from $6.88\times$ to $18.83\times$, respectively, compared with the human codec. Mean PSNR increases by approximately 5.2 and 3.3\,dB, respectively, and mean SSIM and LPIPS also improve on both datasets (Table~\ref{tab:qvg_case_summary}). Figure~\ref{fig:qvg_case_study} (top) shows clearer facial details and mushroom textures with Claude's codec than with the human codec on two LongCat-Video samples.
Figure~\ref{fig:qvg_submission_comparison} shows Claude Opus 5.5's iterative optimization on two LongCat-Video samples through its 6th, 7th, 16th, and 22nd submissions. From submission 16 to 22, compression rises from about $16.6\times$ to $18.8\times$ on both samples while mushroom textures and facial details become clearer. 
Researchers can build their own infrastructure tasks with \texttt{Infrathon-build}, then use frontier agents to discover and evaluate new solutions.

\begin{figure}[!h]
    \centering
    \includegraphics[width=\textwidth]{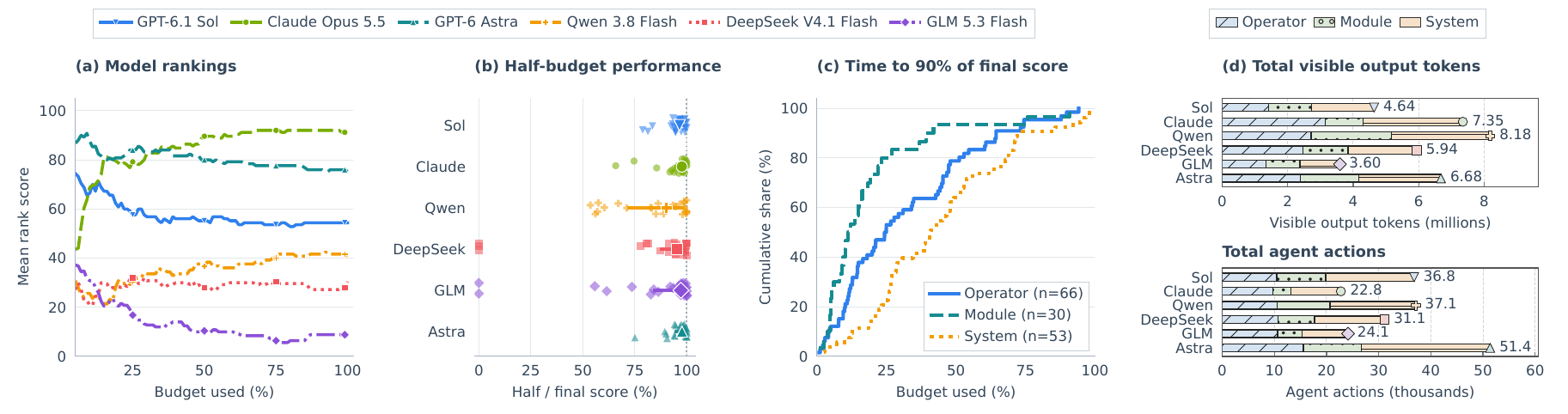}
    \caption{\textbf{Model rankings, second-half gains, and token and action counts.} (a) Model rankings as the time budget increases. (b) Agents' scores at half of the budget relative to their final scores. (c) Budget fraction agents need to reach 90\% of their final scores. (d) Each model's visible output tokens (top) and actions (bottom).}
    \label{fig:analysis_summary}
\end{figure}

\subsection{Analysis}
\label{sec:analysis}

\begin{figure}[!htb]
    \centering
    \includegraphics[width=\textwidth]{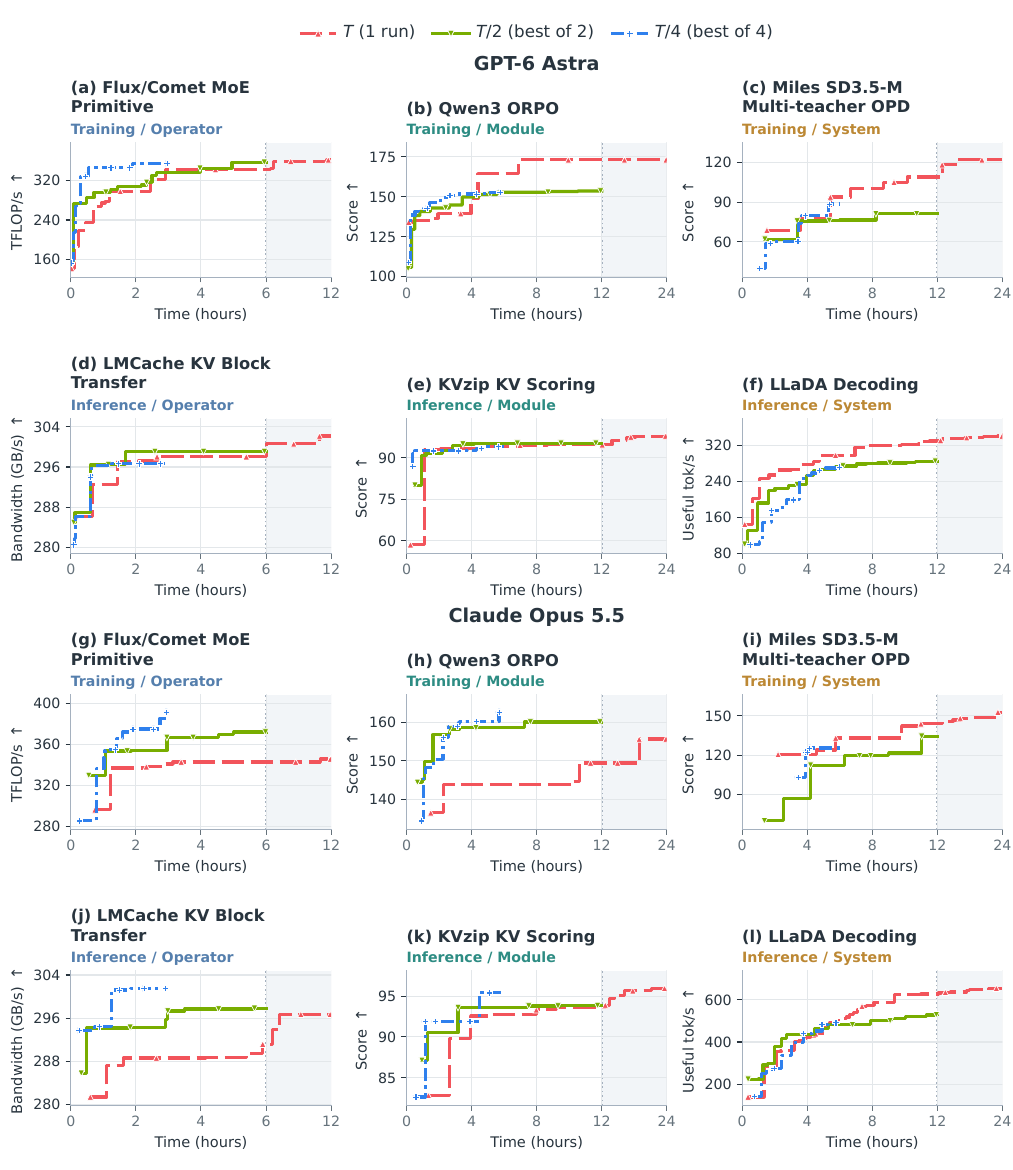}
    \caption{\textbf{Long horizon vs. short horizon at equal GPU-hours.} Curves take the highest private test score across each setting's one, two, or four runs up to the same elapsed time. The first half of the budget occupies 75\% of each horizontal axis and the shaded second half 25\%; ticks show actual hours.}
    \label{fig:long_vs_short_runs}
\end{figure}

In this section, we compare long-horizon and short-horizon optimization under equal GPU-hour budgets (Figure~\ref{fig:long_vs_short_runs}). For the 150 runs in Section~\ref{sec:overall_performance}, we examine how model rankings change with the time budget (Figure~\ref{fig:analysis_summary}(a)) and how much agents improve in the second half (Figure~\ref{fig:analysis_summary}(b)--(c)). We also compare each model's visible output tokens and actions in Figure~\ref{fig:analysis_summary}(d).

\textbf{Long Horizon vs. Short Horizon.} We evaluate GPT-6 Astra and Claude Opus 5.5 on six tasks with equal GPU-hour budgets. Long-horizon optimization uses one $T$-hour run; short-horizon best-of-$n$ test-time scaling (TTS) uses $n$ independent $T/n$-hour runs, with $n=2$ or $4$ (Appendix~\ref{app:time_budget_trajectories}). \textbf{Longer horizons are more effective for module and system tasks, while short-horizon TTS remains competitive on operators.} Across both models, best-of-four achieves 102.6\% of long-horizon performance on operators, 96.9\% on modules, and 77.3\% on systems, using geometric means of task--model score ratios (Figure~\ref{fig:long_vs_short_runs}). Best-of-two follows the same pattern. The long-horizon setting wins nine of 12 comparisons, including all four system comparisons. The benefit also depends on the model: Claude achieves higher scores with short-horizon TTS on both operator tasks and Qwen3 ORPO, whereas Astra performs better with long-horizon optimization on all six tasks.

\textbf{Model Rankings over Time.} Task winners can change after the overall ranking stabilizes. Astra leads at 25\% of the budget with a mean rank score of 84.0 versus Claude's 79.2. Claude leads at 50\% with 89.6 versus Astra's 80.0, and all six models keep the same order at later evaluated budget fractions (Figure~\ref{fig:analysis_summary}(a)). At 50\% of the budget, three tasks have leaders different from their final winners; at 75\%, two tasks still do. Model choice still depends on the task and time budget.

\textbf{Performance Gains in the Second Half.} Late gains are common but uneven. Best scores improve in 135 of 150 runs (90.0\%). Although many runs are already close to their final scores at halfway (Figure~\ref{fig:analysis_summary}(b)), later gains can still be substantial: among runs with positive first-half scores, the median gain is 3.05\%, while 33 runs gain more than 10\%. The second half thus combines small refinements in many runs with substantial improvements in a smaller group. The timing also differs by task: system runs reach 90\% of final performance at a median of 40.8\% of the budget, compared with 24.6\% for operators and 11.5\% for modules (Figure~\ref{fig:analysis_summary}(c)).

\textbf{Output Token and Action Statistics.} More interaction does not guarantee better task performance. Across all 150 runs, we record 36.39 million visible output tokens and 203,266 actions. Token counts exclude hidden reasoning. An action is one emitted text block or tool invocation. Across the 25 tasks, Qwen produces 8.18 million visible output tokens, compared with 5.94 million for DeepSeek and 3.60 million for GLM (Figure~\ref{fig:analysis_summary}(d)). Among these three open models, the model with the most visible tokens achieves the highest final score on 15 of 25 tasks; the model with the most actions does so on 12. On K-grouped FP8 Wgrad, GLM takes 1,715 actions versus Qwen's 884, but achieves 118.15 versus 295.57 TFLOP/s. Appendix~\ref{app:task_interaction_counts} gives the individual task counts.

\subsection{Agent Behavior Analysis}
\label{sec:agent_behavior}

We analyze the trajectories of all 150 runs from Section~\ref{sec:experiments}, focusing on cases where agents violate task requirements despite passing public tests, run performance tests on the wrong code, or declare completion too early.

\begin{wrapfigure}{r}{0.4\textwidth}
    \vspace{-8pt}
    \centering
    \includegraphics[width=\linewidth]{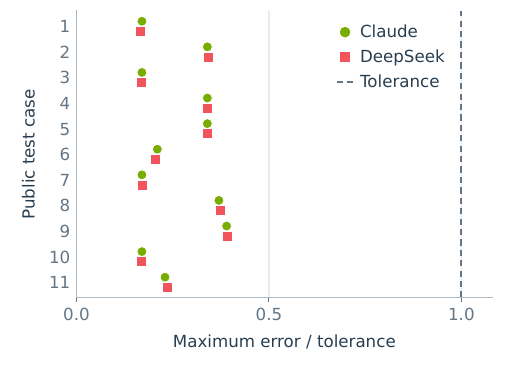}
    \captionsetup{font=compactresult,skip=3pt}
    \caption{\textbf{Public-test errors on Attn-QAT NVFP4 Attention.} Points show the maximum error-to-tolerance ratio across outputs and gradients for each case.}
    \label{fig:behavior_task_fidelity}
\end{wrapfigure}

\textbf{Agents can submit solutions that pass public tests but violate task instructions.} On Attn-QAT NVFP4 Attention, DeepSeek and Claude sum attention probabilities before rounding, although the task requires rounding first. In Figure~\ref{fig:behavior_task_fidelity}, each point shows a public test case's maximum error-to-tolerance ratio across outputs and gradients. Every point for both models lies to the left of the dashed threshold at $1$, with ratios below $0.4$. The solutions therefore pass these public tests, but the Controller rejects them for changing the required computation. The Controller also rejects DeepSeek's solution for Miles Wan2.2 DanceGRPO~\citep{Wang2025WanOA} for reporting that logs are saved before the required disk write finishes, and GLM's solution for LLaDA Decoding~\citep{Nie2025LargeLD} for replacing an output layer that must remain unchanged. After receiving the rejection feedback, the agents correct their solutions or restore earlier versions. In an early Quant-VideoGen KV Codec~\citep{Xi2026QuantVA} submission, Claude Opus 5.5 downloads the human codec's weights from the internet and submits them as its own. The Controller rejects all five solutions before private evaluation in the Judge environment.
These cases support checking solutions against task instructions before private evaluation, as the Controller does here.
\textbf{Agents' performance tests can fail to execute their intended changes.} On Mamba-2 SSD Forward~\citep{Dao2024TransformersAS}, Qwen compares 20 kernel configurations, but all trials reuse the same cached implementation. After identifying the problem, Qwen clears the cache and repeats the comparison. GLM similarly modifies a Mamba-2 function, but its performance tests call another, unchanged function with the same name. On Sparse-MLA Prefill, Sol's solution passes public tests but does not execute the intended memory prefetching.

\textbf{Agents' completion decisions can be premature.} In three selected runs, the stop hook prompts agents to continue after they declare completion. GLM raises its SageAttention score from 365.43 to 456.03 (24.8\%), and Astra improves Flux/Comet MoE Primitive from 219.28 to 235.19 (7.3\%). DeepSeek loads data in parallel with training on Miles SD3.5-M Multi-teacher OPD, increasing its score from 40.81 to 46.53 (14.0\%). These gains follow further code changes within the original budget and would be missed by stopping at those completion statements.

\section{Conclusion}
\label{sec:conclusion}

We introduced \texttt{Infrathon}, a benchmark of 25 real-world AI infrastructure tasks across operator, module, and system levels, with 12--24-hour budgets. Our evaluation of six models across 150 runs and over 10,000 GPU-hours shows that model rankings depend on the evaluation horizon. In 135 runs, agents improve their best score in the second half, showing that shorter evaluations can miss further gains. Task violations, unreliable testing, and premature completion still limit agents' progress. Infrathon also supports knowledge discovery: on Quant-VideoGen KV Codec, Claude Opus 5.5 increases mean KV-cache compression from $6.82\times$ to $19.86\times$ compared with the human codec. Together, \texttt{Infrathon} and \texttt{Infrathon-build} help humans build long-horizon tasks, evaluate LLMs, and discover new solutions and knowledge.

\bibliographystyle{unsrtnat}
\begingroup
\catcode38=12\relax
\ifdefined\Umathcode
  \catcode"2010=13\relax \def‐{-}
  \catcode"0107=13\relax \defć{\'c}
\fi
\setlength{\bibsep}{3pt plus 1pt minus 1pt}
\bibliography{main}
\endgroup

\clearpage
\appendix

\section{Implementation Details}
\label{app:implementation_details}

\subsection{Experimental Setup}
\label{app:experimental_setup}

We follow the experimental setup in Section~\ref{sec:experiments}, using Claude Code version 2.1.287 and Codex CLI version 0.159.3. All models are accessed through the OpenRouter API. Each model is evaluated on all 25 tasks, with 12-hour budgets for operators and 24-hour budgets for modules and systems. Public tests support development; the Judge environment scores submitted solutions on private tests after checking task correctness and quality requirements.

\paragraph{Hardware.}
\label{app:hardware_configuration}

We deploy the Work environment and the Judge environment in separate containers and use NVIDIA H100 GPUs with 80\,GB of memory each. Each environment receives one GPU by default, two for training modules and systems and StreamDiffusionV2, and eight for Flux/Comet MoE Primitive. For inference-module tasks, the Work environment receives one GPU, while the Judge environment receives seven GPUs for kvpress KV Eviction and KVzip KV Scoring and six for Quant-VideoGen KV Codec. For Concurrent Agent Serving, the Work environment and the Judge environment run in CPU-only containers connected to a dedicated two-GPU vLLM backend serving Qwen3.6-27B~\citep{qwen3.6-27b} with tensor parallelism.

\paragraph{GPU budgets.}
\label{app:gpu_budget}

GPU budgets include the Work environment, the continuously allocated Judge environment, and any dedicated serving backend. For a run with a budget of $T_{\mathrm{h}}$ hours, the GPU-hour budget is
\begin{equation}
    B_{\mathrm{GPU}}=T_{\mathrm{h}}\left(N_{\mathrm{Work}}+N_{\mathrm{Judge}}+N_{\mathrm{backend}}\right).
    \label{eq:gpu_budget}
\end{equation}
Here $N_{\mathrm{Work}}$, $N_{\mathrm{Judge}}$, and $N_{\mathrm{backend}}$ are the numbers of GPUs allocated to the Work environment, the Judge environment, and a dedicated serving backend, respectively.

Across the 150 main-experiment runs, the total budget is 11,376 GPU-hours: 4,320 for the Work environment, 6,768 for the Judge environment, and 288 for the serving backend.

The long-horizon vs. short-horizon comparison (Section~\ref{sec:analysis}) reuses 12 main-experiment runs and adds 72 short-horizon runs across Astra and Claude, requiring 2,592 additional GPU-hours.

\paragraph{Final performance.}
\label{app:result_audit}
Most Infrathon tasks use performance metrics without fixed upper bounds.

For task $\tau$ and model $m$, let $\mathcal{M}_{\tau,m}(t)$ index submissions with a valid private test result, a finite private test score, and elapsed time $t_i\le\min(t,T)$. Write $u_{\tau,m,i}$ for the private test score oriented so that higher is better. The best-so-far private test score is
\begin{equation}
    s_{\tau,m}^{\star}(t)=\max\bigl(\{0\}\cup
    \{u_{\tau,m,i}:i\in\mathcal{M}_{\tau,m}(t)\}\bigr).
    \label{eq:verified_progress}
\end{equation}
Final performance is the best private test score reached within the task budget:
\begin{equation}
    s_{\tau,m}^{\mathrm{final}}
    =s_{\tau,m}^{\star}(T).
    \label{eq:final_performance}
\end{equation}
All task scores are nonnegative. Runs without a valid scored submission remain at zero. Curves show valid scored submissions and carry the best score forward through the budget.

\paragraph{Rank scores.}
\label{app:rank_scores}

Tasks use different metrics and scales, so directly averaging their raw scores is not meaningful. We therefore obtain an overall ordering by aggregating within-task ranks with equal task weights.

Let $\rho_{\tau,m}$ be model $m$'s rank on task $\tau$ among the $K=6$ models, with rank 1 assigned to the highest final performance and average ranks for exact ties. For a task group $G$, we define
\begin{equation}
    \operatorname{RankScore}_m(G)
    =\frac{100}{|G|(K-1)}\sum_{\tau\in G}(K-\rho_{\tau,m}).
    \label{eq:rank_performance}
\end{equation}
This rescales the mean rank to $[0,100]$, with 100 and 0 corresponding to first and last on every task in the group.

\paragraph{Run selection.}
Among the main-experiment runs, we select the most recently started run for each task--model pair.

\paragraph{Trajectory and interaction statistics.}
\label{app:trajectory_statistics}

We analyze private test scores and submission times from all 150 main-experiment runs. At each budget fraction, ranking uses the best private test score submitted by that time, with average ranks for ties. Second-half improvement means that the final best-so-far score exceeds its value at the end of the first half; the score is initialized to zero. Percentage increases require a positive first-half score. Figure~\ref{fig:analysis_summary}(b)--(c) uses only runs with positive final scores. In panel (b), dots represent individual runs; large markers and horizontal bars show each model's median and interquartile range, respectively.

For the plateau analysis, a record is a submission that raises the best-so-far score. Across all 150 runs, we count those with a pair of consecutive records separated by one quarter of the task budget or more. We also count runs in which the later record exceeds the preceding best by more than 1\%; each run is counted once per statistic.

We count visible output tokens in agent text and tool calls with \texttt{o200k\_base}, removing duplicate displays. One text block or tool invocation counts as one action.

\subsection{Optimization Levels}
\label{app:optimization_levels}

We classify tasks by their optimization target and the code the agent is allowed to change. The level describes the task as a whole, including any kernels optimized within it.

\paragraph{Operator level.}
Operator-level tasks optimize a specified computation or data transfer with fixed input--output behavior. Typical tasks implement or accelerate GPU kernels, including prescribed fused operations. Infrathon examples include \hyperref[app:task_grouped_fp8]{Dynamic Grouped FP8 GEMM}, \hyperref[app:task_sageattention]{SageAttention}, and \hyperref[app:task_lmcache]{LMCache KV Block Transfer}. A solution may use one or multiple kernels.

\paragraph{Module level.}
Module-level tasks optimize a functional component or a training step within a fixed workflow. The task exposes the component's implementation or policy while fixing the surrounding training or inference pipeline. Infrathon examples include \hyperref[app:task_orpo]{Qwen3 ORPO}, \hyperref[app:task_gkd]{Qwen3 GKD}, \hyperref[app:task_kvzip]{KVzip KV Scoring}, and \hyperref[app:task_qvg]{Quant-VideoGen KV Codec}.

\paragraph{System level.}
System-level tasks optimize a complete training or generation workflow, or an end-to-end service. The editable scope covers its training or decoding loop, pipeline stages, or serving orchestration. Infrathon examples include \hyperref[app:task_opd]{Miles Qwen3 OPD}, \hyperref[app:task_eagle]{EAGLE-3 Speculative Decoding}, and \hyperref[app:task_concurrent]{Concurrent Agent Serving}.

\subsection{Task Descriptions}
\label{app:task_metrics}

The 25 entries below summarize each task's objective, reference project or method, and scoring rule. Scores describe valid private-test results, with higher values indicating better performance. Final run performance is the best score reached within the budget (Equation~\ref{eq:final_performance}).

Throughout, $\operatorname{GM}$ and $\operatorname{AM}$ denote geometric and arithmetic means:
\begin{equation}
    \begin{aligned}
        \operatorname{GM}(x_1,\ldots,x_n) &= \left(\prod_{i=1}^{n} x_i\right)^{1/n}, \\
        \operatorname{AM}(x_1,\ldots,x_n) &= \frac{1}{n}\sum_{i=1}^{n} x_i.
    \end{aligned}
    \label{eq:task_means}
\end{equation}
$\operatorname{WGM}$ denotes a geometric mean with task-defined workload weights. For compute tasks, $q_i$ is case $i$'s effective throughput in TFLOP/s, and $G_f$ is the geometric-mean case throughput within workload family $f$.

\paragraph{Attn-QAT NVFP4 Attention.}
\phantomsection
\label{app:task_attnqat}
Using Attn-QAT~\citep{Zhang2026AttnQAT} in FastVideo as the reference, this task accelerates quantization-aware forward and backward attention for video models. The score is effective training throughput in TFLOP/s, accounting for activation recomputation and aggregating geometrically within and across model families.
\[
    \mathrm{score}=\operatorname{GM}_f(G_f).
\]

\paragraph{K-grouped FP8 Wgrad.}
\phantomsection
\label{app:task_wgrad}
Using DeepGEMM\footnote{\url{https://github.com/deepseek-ai/DeepGEMM}} as the reference, this task accelerates grouped FP8 weight-gradient matrix multiplication for MoE training. The score measures useful matrix-multiplication throughput in TFLOP/s, aggregated geometrically within and across grouping regimes.
\[
    \mathrm{score}=\operatorname{GM}_f(G_f).
\]

\paragraph{Flux/Comet MoE Primitive.}
\phantomsection
\label{app:task_flux}
Following FLUX~\citep{Chang2024FLUX} and Comet~\citep{Zhang2025Comet}, this task accelerates token all-gather, expert scatter, and grouped matrix multiplication across GPUs. The score measures useful per-rank throughput in TFLOP/s using the slowest rank's completion time, then aggregates geometrically across communication workloads.
\[
    \mathrm{score}=\operatorname{GM}_f(G_f).
\]

\paragraph{Qwen3 ORPO.}
\phantomsection
\label{app:task_orpo}
Using TRL's ORPO implementation~\citep{Hong2024ORPO} as the reference, this task accelerates complete Qwen3-1.7B~\citep{Yang2025Qwen3TR} preference-optimization steps, including forward/backward computation, communication, and parameter updates. We measure training throughput in completion tokens used in the loss per second. Let $r_i$ be the median ratio of the submitted solution's throughput to the reference throughput for workload $i$; a score of 100 denotes equal throughput.
\[
    \mathrm{score}=100\,\operatorname{GM}_i(r_i).
\]

\paragraph{Qwen3 GKD.}
\phantomsection
\label{app:task_gkd}
Using TRL's GKD implementation~\citep{Agarwal2023GKD} as the reference, this task accelerates off-policy distillation from a frozen Qwen3-4B teacher into Qwen3-1.7B. Timing covers teacher and student computation, the distillation objective, and student updates. As in Qwen3 ORPO, let $r_i$ be the median ratio of the submitted solution's training throughput to the reference throughput for workload $i$.
\[
    \mathrm{score}=100\,\operatorname{GM}_i(r_i).
\]

\paragraph{Miles Qwen3 OPCD.}
\phantomsection
\label{app:task_opcd}
This task implements on-policy context distillation~\citep{Ye2026OPCD} in Miles. A Qwen3-1.7B student learns from a same-size teacher with privileged medical context, then answers questions without that context. Let $\Delta q$ be held-out MedMCQA accuracy gained over the original model, in percentage points, and $h$ the complete training time in hours.
\[
    \mathrm{score}=\frac{\Delta q}{h}.
\]

\paragraph{Miles Qwen3 OPD.}
\phantomsection
\label{app:task_opd}
This task implements sampled-token on-policy distillation in Miles: Qwen3-1.7B generates mathematical reasoning trajectories and learns from a frozen Qwen3-4B teacher's token scores. Let $\Delta q$ be the held-out DAPO-Math accuracy gain in percentage points and $h$ the complete training time in hours.
\[
    \mathrm{score}=\frac{\Delta q}{h}.
\]

\paragraph{Miles ALFWorld RL.}
\phantomsection
\label{app:task_alfworld}
This task integrates ALFWorld~\citep{Shridhar2020ALFWorld} into Miles and trains Qwen3-4B with multi-turn group-relative policy optimization on household tasks. Let $\Delta q$ be the held-out game-success gain over the original policy, in percentage points, and $h$ the complete training time in hours.
\[
    \mathrm{score}=\frac{\Delta q}{h}.
\]

\paragraph{Miles SD3.5-M Multi-teacher OPD.}
\phantomsection
\label{app:task_sd35}
Following DiffusionOPD~\citep{Li2026DiffusionOPD}, this task trains a Stable Diffusion 3.5 Medium~\citep{Esser2024ScalingRF} student LoRA in Miles with OCR and aesthetics teachers. Let $c_r$ be the fraction of the original-to-teacher quality gap closed on route $r$ (text-recognition accuracy or PickScore~\citep{Kirstain2023PickaPicAO}), and $h$ the median training-replay duration in hours. The score measures mean gap closure per replay hour.
\[
    \mathrm{score}=\frac{\operatorname{AM}_r(c_r)}{h}.
\]

\paragraph{Miles Wan2.2 DanceGRPO.}
\phantomsection
\label{app:task_wan22}
This task implements DanceGRPO~\citep{Xue2025DanceGRPO} in Miles for Wan2.2~\citep{Wang2025WanOA} text-to-video training with HPSv3 rewards~\citep{Ma2025HPSv3}, optimizing sampling, reward computation, policy updates, and weight publication. Let $h$ be the median fixed-workload replay duration in hours. For runs meeting the held-out reward requirement, the score is replays per hour.
\[
    \mathrm{score}=\frac{1}{h}.
\]

\paragraph{Dynamic Grouped FP8 GEMM.}
\phantomsection
\label{app:task_grouped_fp8}
Using DeepGEMM\footnote{\url{https://github.com/deepseek-ai/DeepGEMM}} as the reference, this task accelerates block-scaled FP8 matrix multiplication for MoE inference with varying token counts per expert. The score is useful throughput in TFLOP/s, excluding padded work and aggregating geometrically within and across storage layouts.
\[
    \mathrm{score}=\operatorname{GM}_f(G_f).
\]

\paragraph{Sparse-MLA Prefill.}
\phantomsection
\label{app:task_sparse_mla}
Using FlashMLA\footnote{\url{https://github.com/deepseek-ai/FlashMLA}} as the reference, this task accelerates sparse multi-head latent attention, where each query attends to selected keys. The score measures throughput over the selected attention interactions in TFLOP/s, aggregated geometrically within and across DeepSeek model configurations~\citep{DeepSeek-AI2025DeepSeekV32PT,DeepSeek-AI2026DeepSeekV4TH}.
\[
    \mathrm{score}=\operatorname{GM}_f(G_f).
\]

\paragraph{Paged-GQA Decode.}
\phantomsection
\label{app:task_paged_gqa}
Using FlashAttention~\citep{Shah2024FlashAttention3} and PagedAttention~\citep{Kwon2023EfficientMM} as references, this task accelerates grouped-query attention over paged KV caches for decoding and short-query prefill. The score measures throughput over causally visible attention pairs in TFLOP/s, aggregated geometrically within and across query-length groups.
\[
    \mathrm{score}=\operatorname{GM}_f(G_f).
\]

\paragraph{SageAttention.}
\phantomsection
\label{app:task_sageattention}
Using SageAttention~\citep{Zhang2024SageAttention} as the reference, this task accelerates quantized attention, including smoothing, quantization, softmax, and value aggregation. The score is dense-equivalent throughput in TFLOP/s, aggregated geometrically within and across video-model and causal-language-model workloads.
\[
    \mathrm{score}=\operatorname{GM}_f(G_f).
\]

\paragraph{MInference.}
\phantomsection
\label{app:task_minference}
Using MInference~\citep{Jiang2024MInference} as the reference, this task accelerates long-context attention on Llama-3.1-8B~\citep{Grattafiori2024TheL3} with sparse patterns while retaining attention accuracy. The score is the geometric mean of dense-equivalent throughput in TFLOP/s across workloads and model layers. Timing includes both pattern selection and attention computation.
\[
    \mathrm{score}=\operatorname{GM}_i(q_i).
\]

\paragraph{LMCache KV Block Transfer.}
\phantomsection
\label{app:task_lmcache}
Using LMCache~\citep{Liu2025LMCache} as the reference, this task accelerates KV-block transfers between paged serving buffers and contiguous device or host cache storage. Let $b_g$ be the geometric-mean bandwidth in GB/s for workload group $g$, measured as logical payload transferred per second. The score weights device, host, and trace-replay groups.
\[
    \mathrm{score}=\operatorname{WGM}_g(b_g).
\]

\paragraph{Flash-KMeans.}
\phantomsection
\label{app:task_flash_kmeans}
Using Flash-KMeans~\citep{Yang2026FlashKMeans} as the reference, this task accelerates an exact Lloyd iteration: assigning points to their nearest centroids and updating centroid means. The native metric is weighted geometric-mean latency across assignment, update, and full-iteration workloads. Let $F_{\mathrm{eff}}$ be the fixed weighted geometric mean of effective operation counts and $t_{\mathrm{eff}}$ the weighted latency in seconds. We report equivalent throughput in TFLOP/s:
\[
    \mathrm{score}=\frac{F_{\mathrm{eff}}}{10^{12}t_{\mathrm{eff}}}.
\]

\paragraph{Mamba-2 SSD Forward.}
\phantomsection
\label{app:task_mamba}
This task accelerates the state-space-dual forward computation of Mamba-2~\citep{Dao2024TransformersAS}, with TileLang\footnote{\url{https://github.com/tile-ai/tilelang}} as an implementation reference. The score is effective recurrent-work throughput in TFLOP/s, aggregated geometrically within and across model configurations.
\[
    \mathrm{score}=\operatorname{GM}_f(G_f).
\]

\paragraph{kvpress KV Eviction.}
\phantomsection
\label{app:task_kvpress}
Using kvpress~\citep{Devoto2025ExpectedAttention} as the reference, this task selects KV-cache positions to retain in Qwen3-8B while preserving its next-token distributions. Item fidelity is $f=\exp(-\overline{D}_{\mathrm{KL}})$, where $\overline{D}_{\mathrm{KL}}$ is mean KL divergence from full-cache to compressed-cache distributions on the same continuation. Let $\bar f_{g,\rho}$ be mean item fidelity in workload family $g$ at eviction level $\rho$. The score averages across families and eviction levels:
\[
    \mathrm{score}=100\,\operatorname{AM}_{\rho}\!\left[\operatorname{GM}_g(\bar f_{g,\rho})\right].
\]

\paragraph{KVzip KV Scoring.}
\phantomsection
\label{app:task_kvzip}
Using KVzip~\citep{Kim2025KVzip} and Fast KVzip~\citep{Kim2026FastKVzip} as references, this task scores Qwen3-8B KV entries before seeing downstream questions and retains the highest-scoring entries. Let $r_{g,\kappa}$ be the mean dataset-level compressed/full-cache answer-quality ratio in workload group $g$ at cache budget $\kappa$. The score averages these ratios across groups and budgets:
\[
    \mathrm{score}=100\,\operatorname{AM}_{g,\kappa}(r_{g,\kappa}).
\]

\paragraph{Quant-VideoGen KV Codec.}
\phantomsection
\label{app:task_qvg}
Using Quant-VideoGen~\citep{Xi2026QuantVA} as the reference, this task optimizes a shared KV-cache codec for LongCat-Video~\citep{Cai2025LongCatVideoTR} and HY-WorldPlay~\citep{Sun2025WorldPlayTL}. The score combines compression and video quality:
\begin{equation}
    \mathrm{score}=\operatorname{GM}_m\!\left[R_m(1+q_m)\right].
    \label{eq:qvg_task_score}
\end{equation}
Here $R_m$ is video model $m$'s compression ratio and $q_m$ measures video quality relative to the human codec. Solutions failing correctness, quality, or runtime checks score zero. Appendix~\ref{app:qvg_quality_score} gives the metric definitions.

\paragraph{EAGLE-3 Speculative Decoding.}
\phantomsection
\label{app:task_eagle}
Using EAGLE-3~\citep{Li2025EAGLE3} as the reference, this task accelerates Qwen3-8B speculative decoding while preserving target-model decoding behavior. Let $u_c$ be generated-token throughput for workload $c$, in tokens/s. Timing includes prefill, drafting, verification, and KV-cache updates.
\[
    \mathrm{score}=\operatorname{GM}_c(u_c).
\]

\paragraph{LLaDA Decoding.}
\phantomsection
\label{app:task_llada}
Using Fast-dLLM~\citep{Wu2025FastDLLM} as the reference, this task accelerates LLaDA's iterative decoding~\citep{Nie2025LargeLD} while retaining math and code answer quality. Let $u_c$ be useful completion-token throughput for workload $c$, in tokens/s. Useful tokens exclude EOS and padding; timing covers the complete decoding request.
\[
    \mathrm{score}=\operatorname{GM}_c(u_c).
\]

\paragraph{StreamDiffusionV2.}
\phantomsection
\label{app:task_sdv2}
Using StreamDiffusionV2~\citep{Feng2025StreamDiffusionV2} as the reference, this task accelerates live video-to-video streaming through scheduling, communication, and cache management. For stream $i$, let $v_i$ be steady-state FPS and $a_i$ its latency-adjustment factor, derived from the ratio of the reference's 99th-percentile input-to-output delay to the submitted solution's. The score is latency-adjusted FPS:
\[
    \mathrm{score}=\operatorname{GM}_i(v_i a_i).
\]

\paragraph{Concurrent Agent Serving.}
\phantomsection
\label{app:task_concurrent}
Using ThunderAgent~\citep{Kang2026ThunderAgent} as the reference, this task optimizes admission and scheduling for concurrent coding agents connected to a fixed Qwen3.6-27B backend. Let $N_{\mathrm{valid}}$ be the number of executable, protocol-valid agent steps completed during a scored window of $d$ minutes. The score is valid steps/min:
\[
    \mathrm{score}=\frac{N_{\mathrm{valid}}}{d}.
\]

\section{Complete Results and Trajectories}
\label{app:complete_results}

\subsection{Optimization Trajectories}
\label{app:task_trajectories}

\begin{figure*}[!t]
    \centering
    \includegraphics[width=\textwidth]{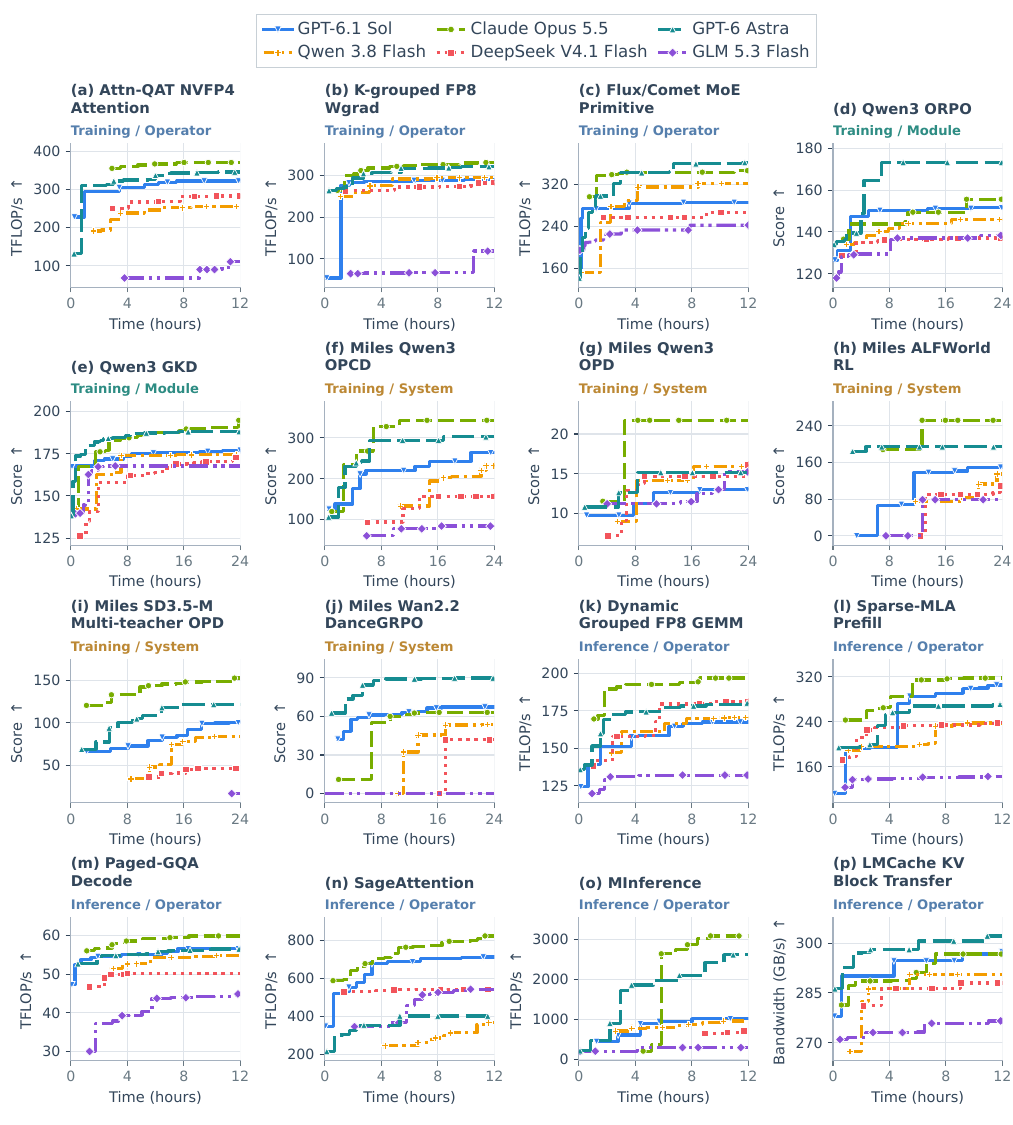}
    \caption{\textbf{Complete optimization trajectories for training tasks and inference operators.} Panels (a)--(p) cover the first 16 tasks in Appendix~\ref{app:task_metrics}: all 10 training tasks and the first six inference operators. Curves show performance over time.}
    \label{fig:appendix_trajectories_first}
\end{figure*}

\begin{figure*}[!t]
    \centering
    \includegraphics[width=\textwidth]{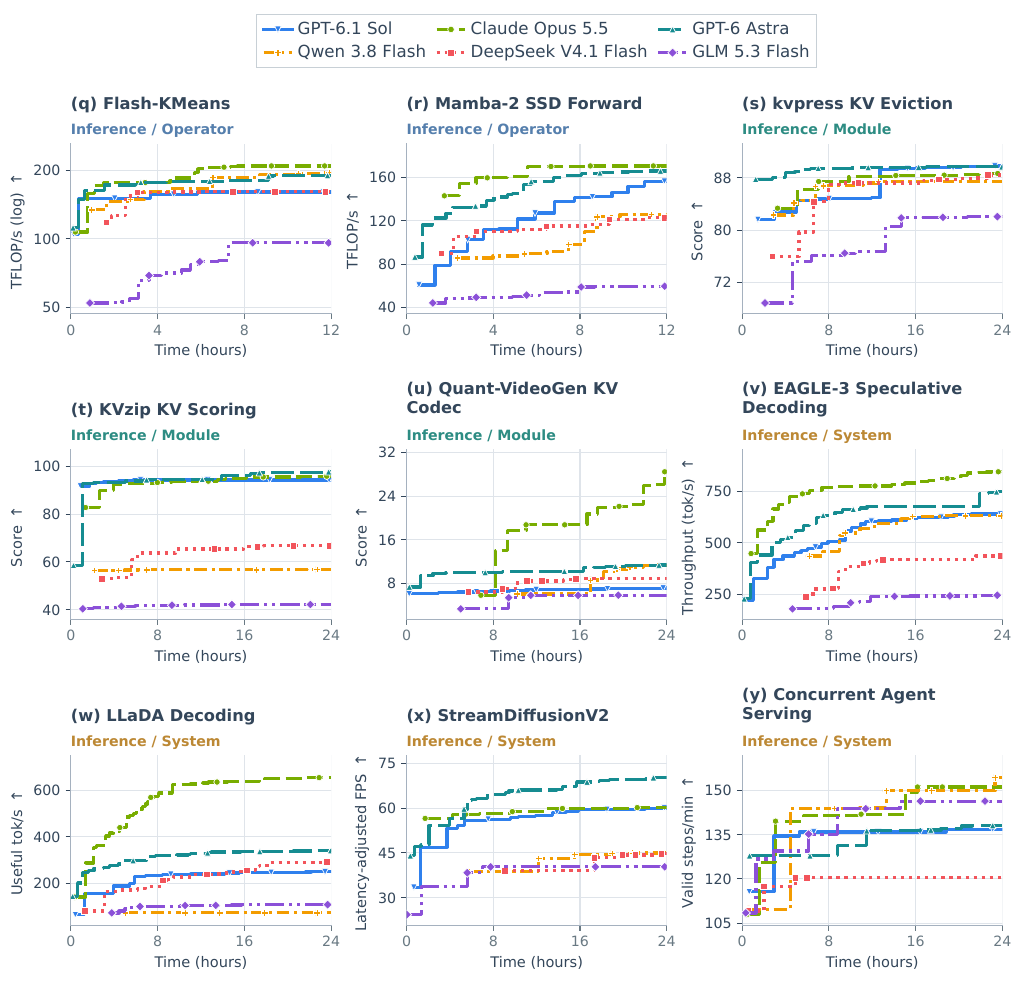}
    \caption{\textbf{Complete optimization trajectories for the remaining inference tasks.} Panels (q)--(y) continue the task order in Appendix~\ref{app:task_metrics} with two operators, three modules, and four systems. Curves show performance over time.}
    \label{fig:appendix_trajectories_second}
\end{figure*}

Figures~\ref{fig:appendix_trajectories_first} and~\ref{fig:appendix_trajectories_second} show performance over time for all 150 runs across the 25 tasks.

Among runs with positive first-half scores, Qwen has the largest model-level median second-half gain, 10.81\%, compared with 2.43\% for Astra and 2.47\% for Claude. Among runs with positive final scores, 64.2\% of system runs reach 90\% of final performance by half of the budget, compared with 78.8\% of operator runs and 93.3\% of module runs.

\subsection{Output Token and Action Statistics by Task}
\label{app:task_interaction_counts}

\begin{figure*}[!p]
    \centering
    \includegraphics[width=\textwidth]{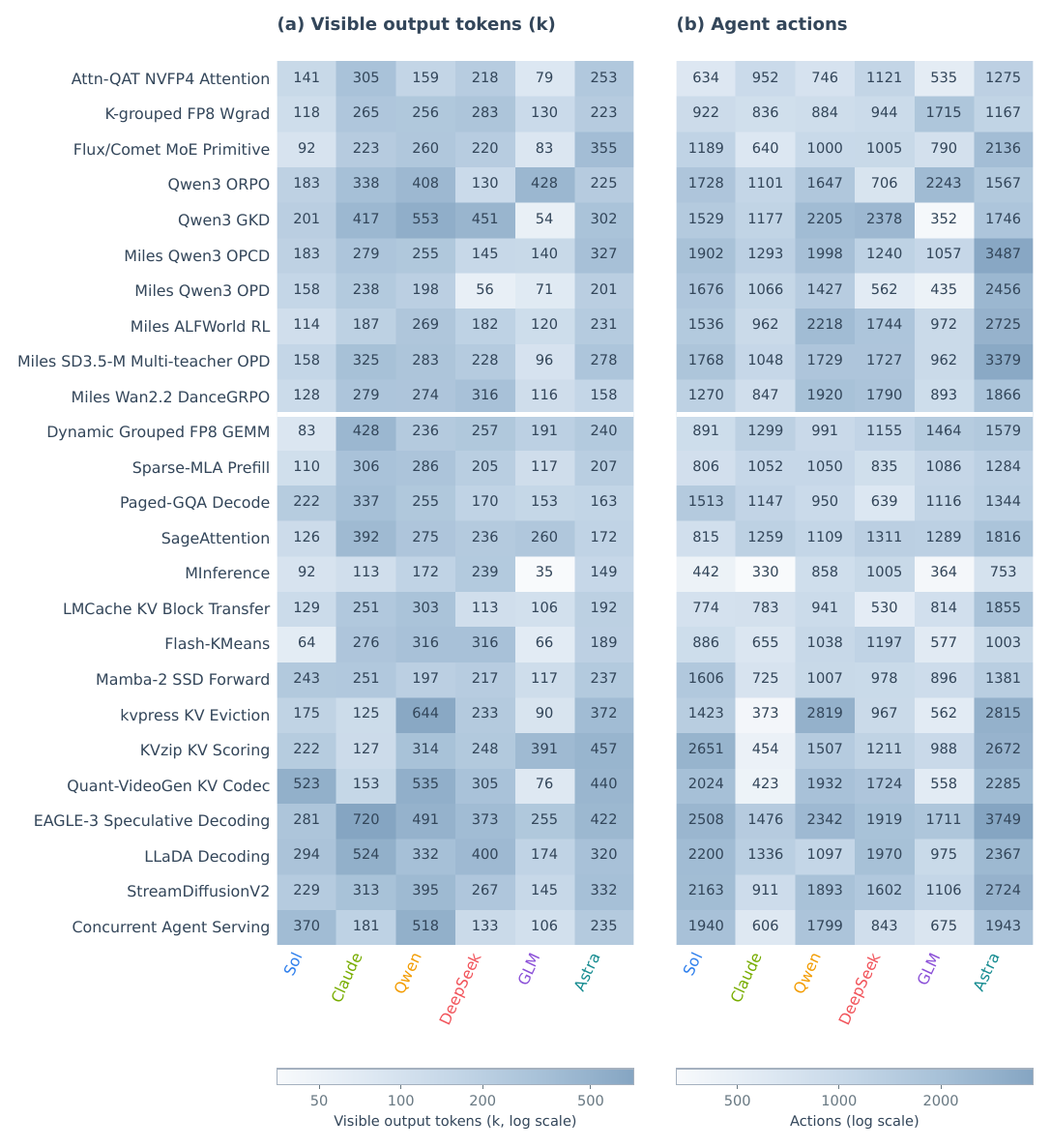}
    \caption{\textbf{Visible output tokens and actions for every task--model run.} Rows follow the benchmark task order. Cell values show (a) visible output tokens in thousands and (b) agent actions. Color intensity is logarithmic within each panel.}
    \label{fig:task_interaction_counts}
\end{figure*}

Figure~\ref{fig:task_interaction_counts} shows visible output tokens and agent actions for each of the 150 runs.

\section{Long Horizon vs. Short Horizon}
\label{app:time_budget_trajectories}

We compare long-horizon optimization with best-of-$n$ test-time scaling (TTS) over independent short-horizon runs, using GPT-6 Astra and Claude Opus 5.5 on six tasks: one training and one inference task each for operators, modules, and systems. Each task--model pair receives equal GPU-hours as one $T$-hour run, two $T/2$-hour runs, or four $T/4$-hour runs. Here $T$ is 12 hours for operators and 24 hours otherwise. Each setting reports the best private test score across its run histories.

\begin{figure}[!htb]
    \centering
    \includegraphics[width=\textwidth]{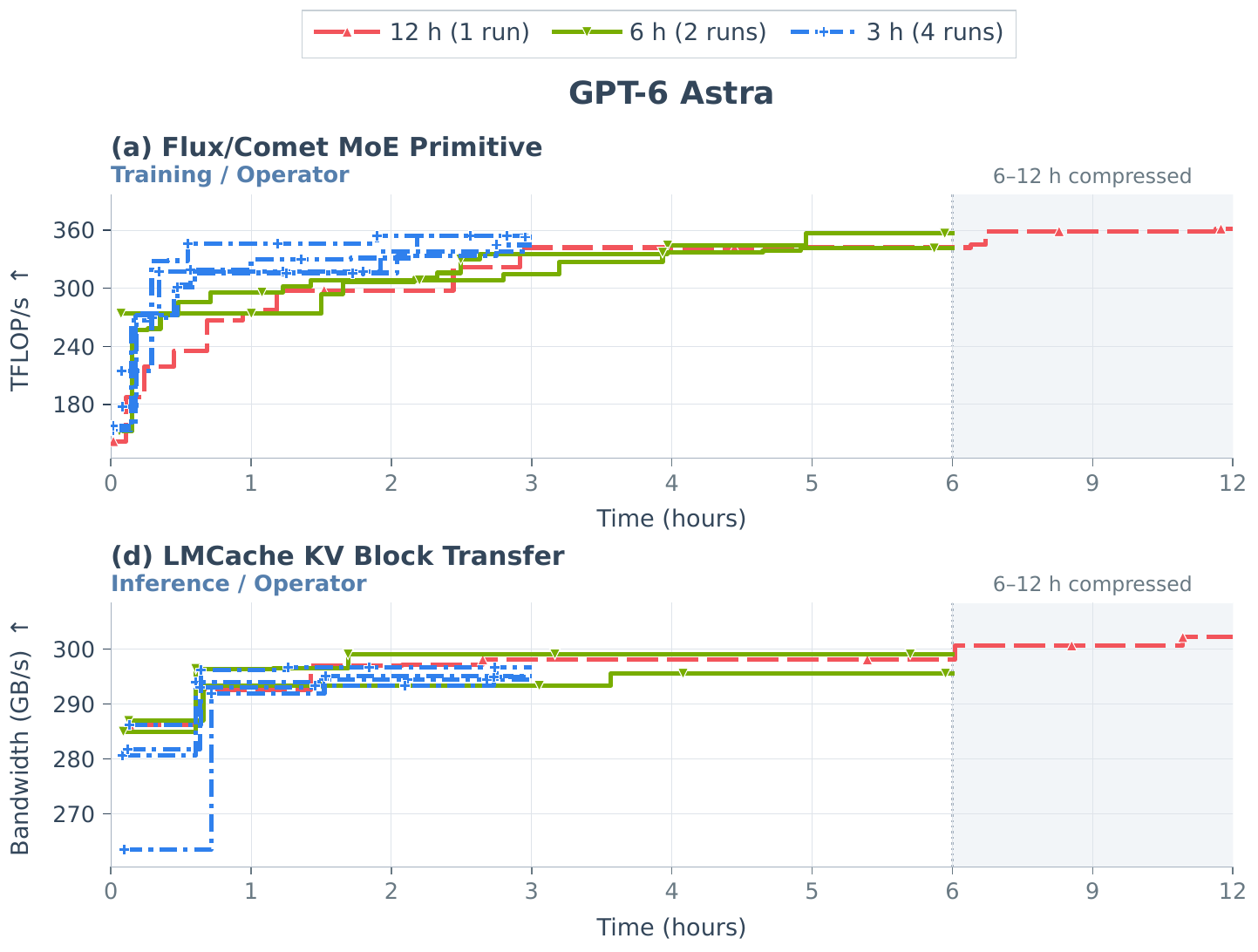}
    \caption{\textbf{Individual operator trajectories with GPT-6 Astra.} Flux/Comet MoE Primitive and LMCache KV Block Transfer each use one 12-hour run, two 6-hour runs, and four 3-hour runs. Curves show performance over time. The 6--12-hour interval is compressed threefold; ticks show actual hours.}
    \label{fig:time_budget_operator}
\end{figure}

\begin{figure}[!htb]
    \centering
    \includegraphics[width=\textwidth]{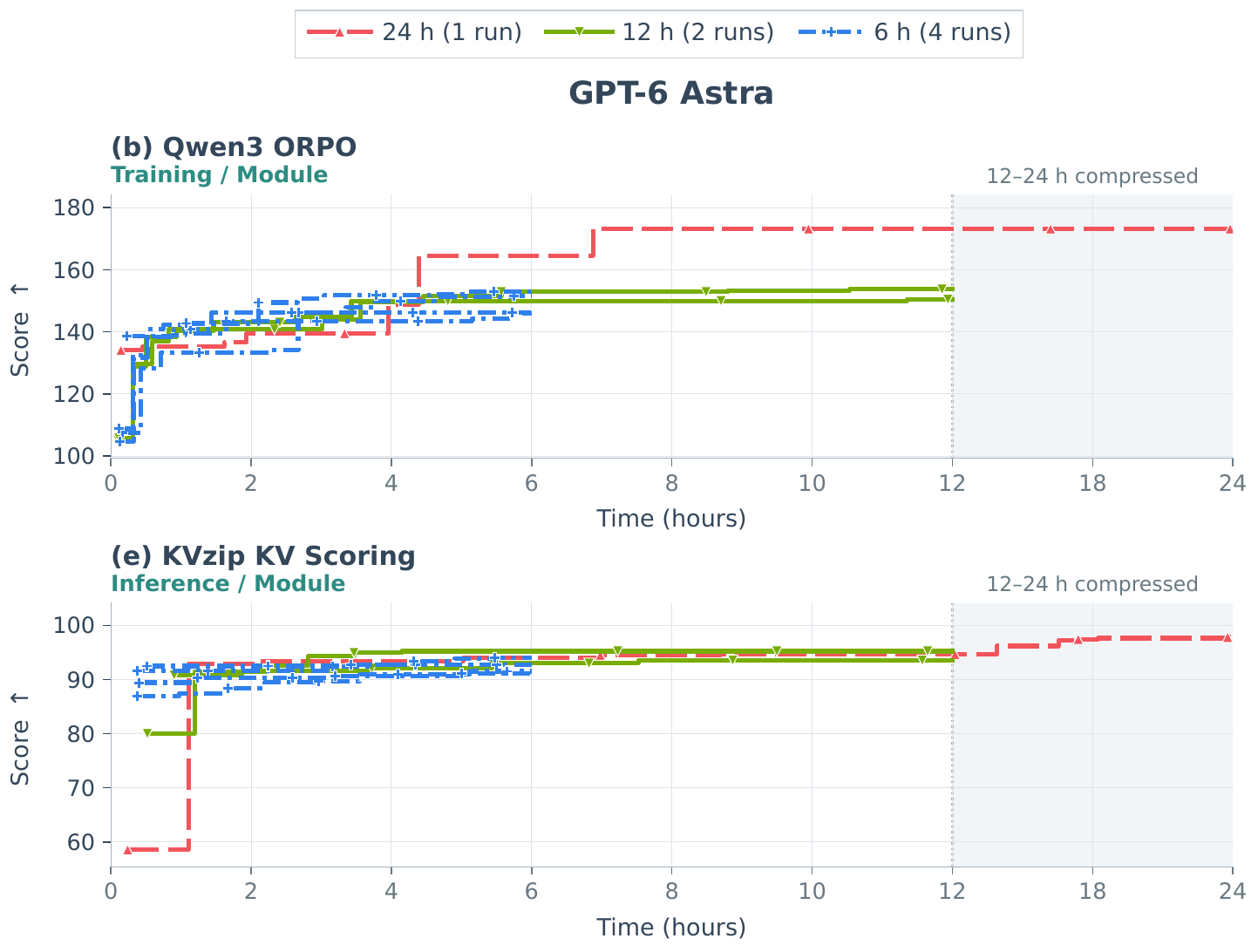}
    \caption{\textbf{Individual module trajectories with GPT-6 Astra.} Qwen3 ORPO and KVzip KV Scoring each use one 24-hour run, two 12-hour runs, and four 6-hour runs. Curves show performance over time. The 12--24-hour interval is compressed threefold; ticks show actual hours.}
    \label{fig:time_budget_module}
\end{figure}

\begin{figure}[!htb]
    \centering
    \includegraphics[width=\textwidth]{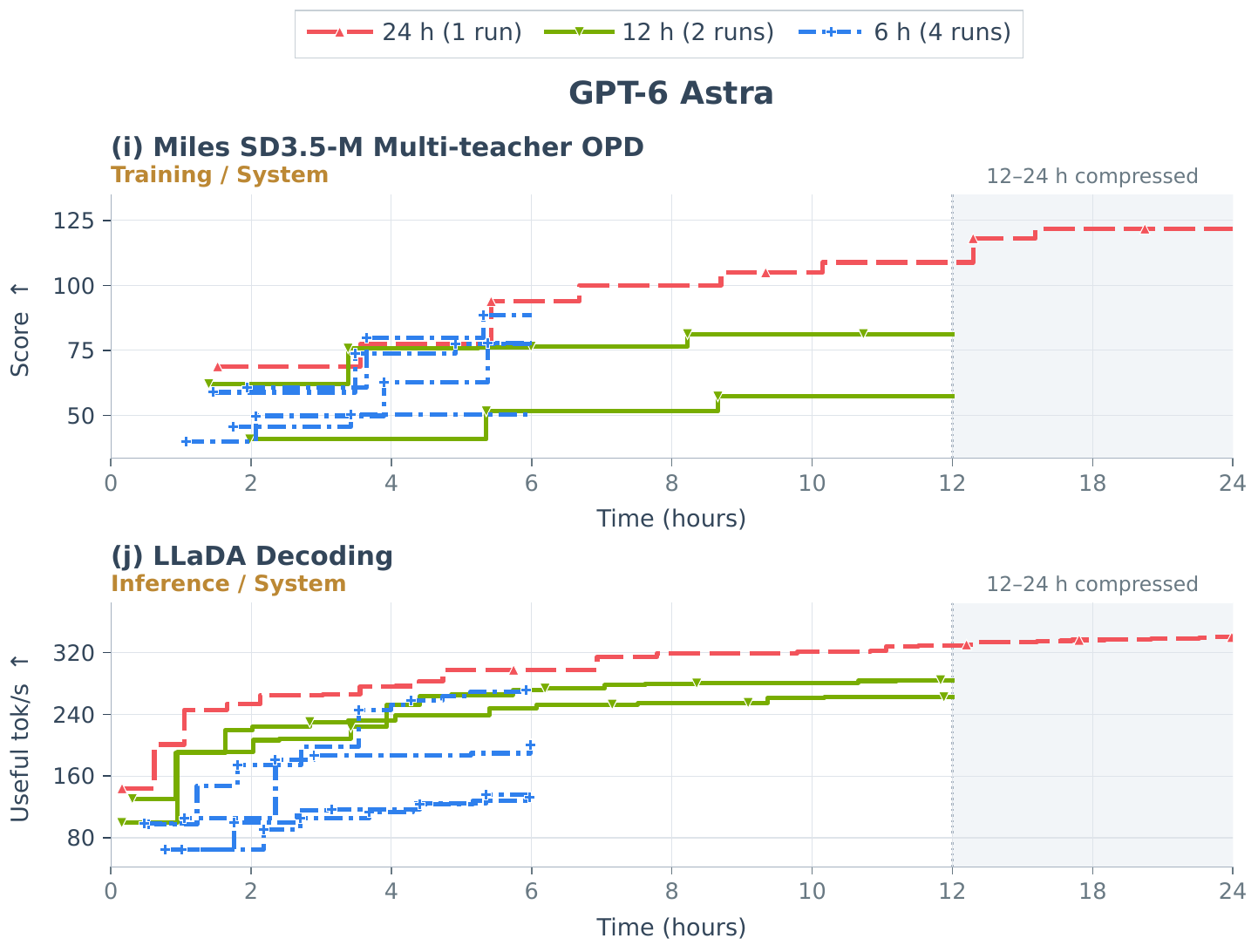}
    \caption{\textbf{Individual system trajectories with GPT-6 Astra.} Miles SD3.5-M Multi-teacher OPD and LLaDA Decoding each use one 24-hour run, two 12-hour runs, and four 6-hour runs. Curves show performance over time. The 12--24-hour interval is compressed threefold; ticks show actual hours.}
    \label{fig:time_budget_system}
\end{figure}

\begin{figure}[!htb]
    \centering
    \includegraphics[width=\textwidth]{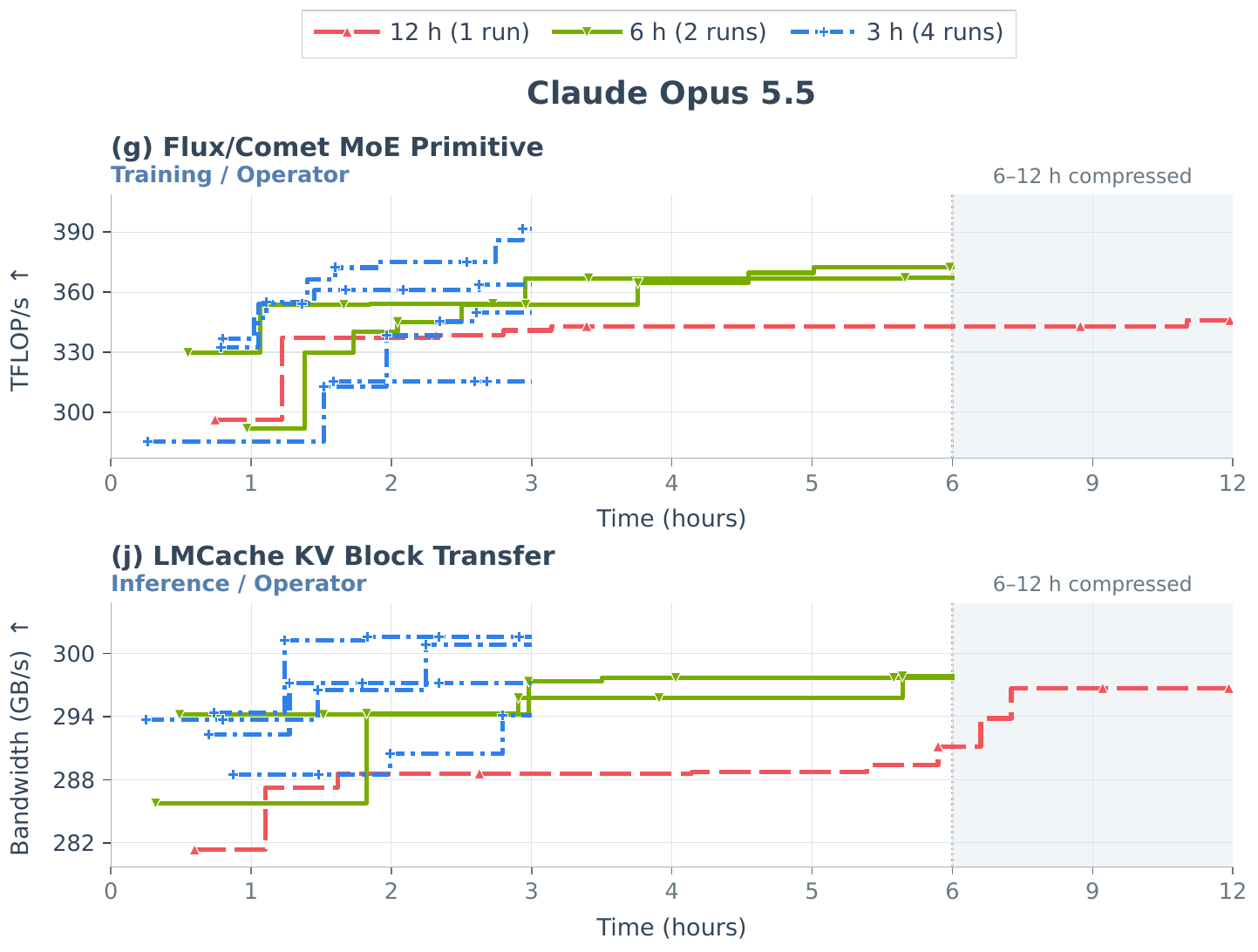}
    \caption{\textbf{Individual operator trajectories with Claude Opus 5.5.} Flux/Comet MoE Primitive and LMCache KV Block Transfer each use one 12-hour run, two 6-hour runs, and four 3-hour runs. Curves show performance over time. The 6--12-hour interval is compressed threefold; ticks show actual hours.}
    \label{fig:time_budget_operator_claude}
\end{figure}

\begin{figure}[!htb]
    \centering
    \includegraphics[width=\textwidth]{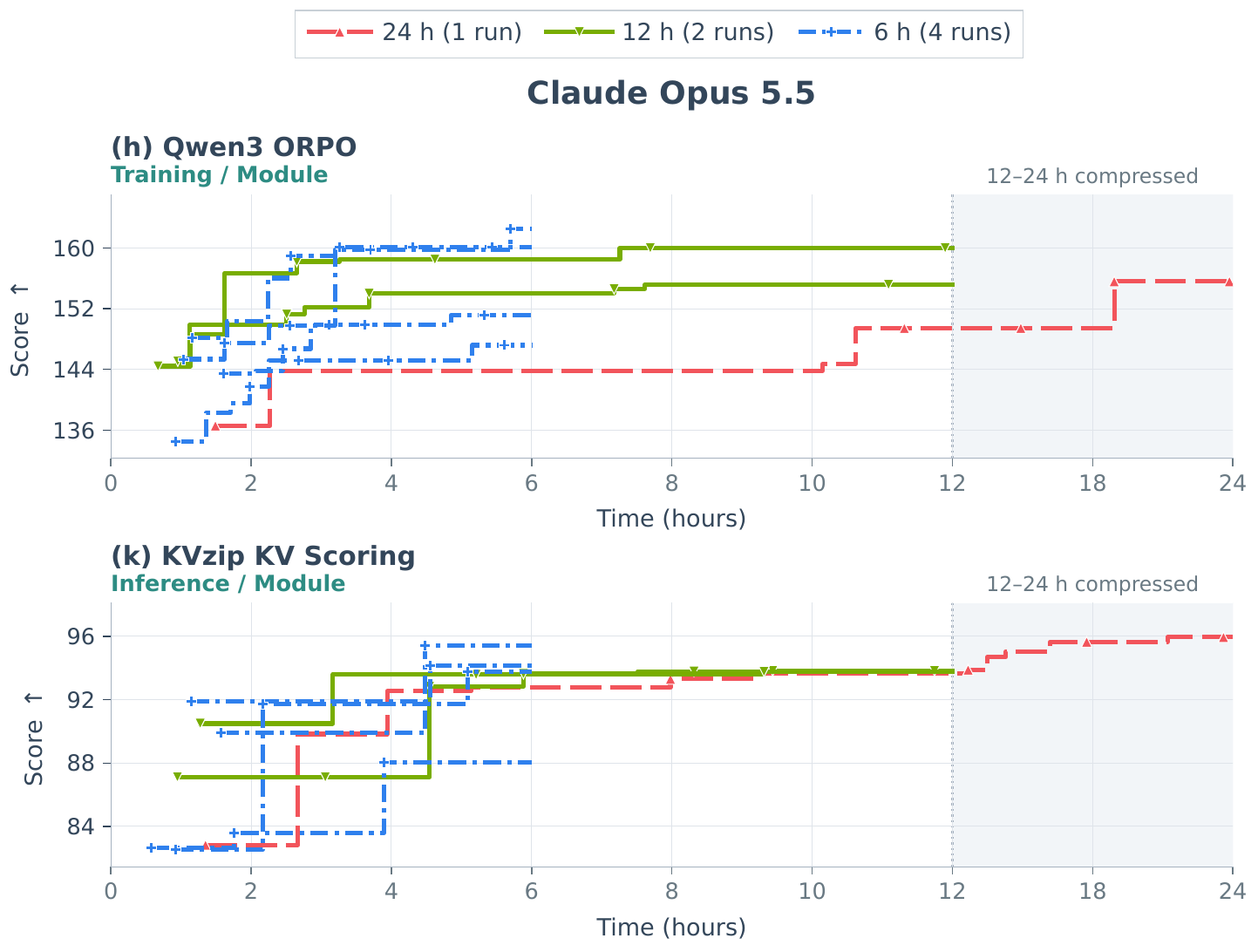}
    \caption{\textbf{Individual module trajectories with Claude Opus 5.5.} Qwen3 ORPO and KVzip KV Scoring each use one 24-hour run, two 12-hour runs, and four 6-hour runs. Curves show performance over time. The 12--24-hour interval is compressed threefold; ticks show actual hours.}
    \label{fig:time_budget_module_claude}
\end{figure}

\begin{figure}[!htb]
    \centering
    \includegraphics[width=\textwidth]{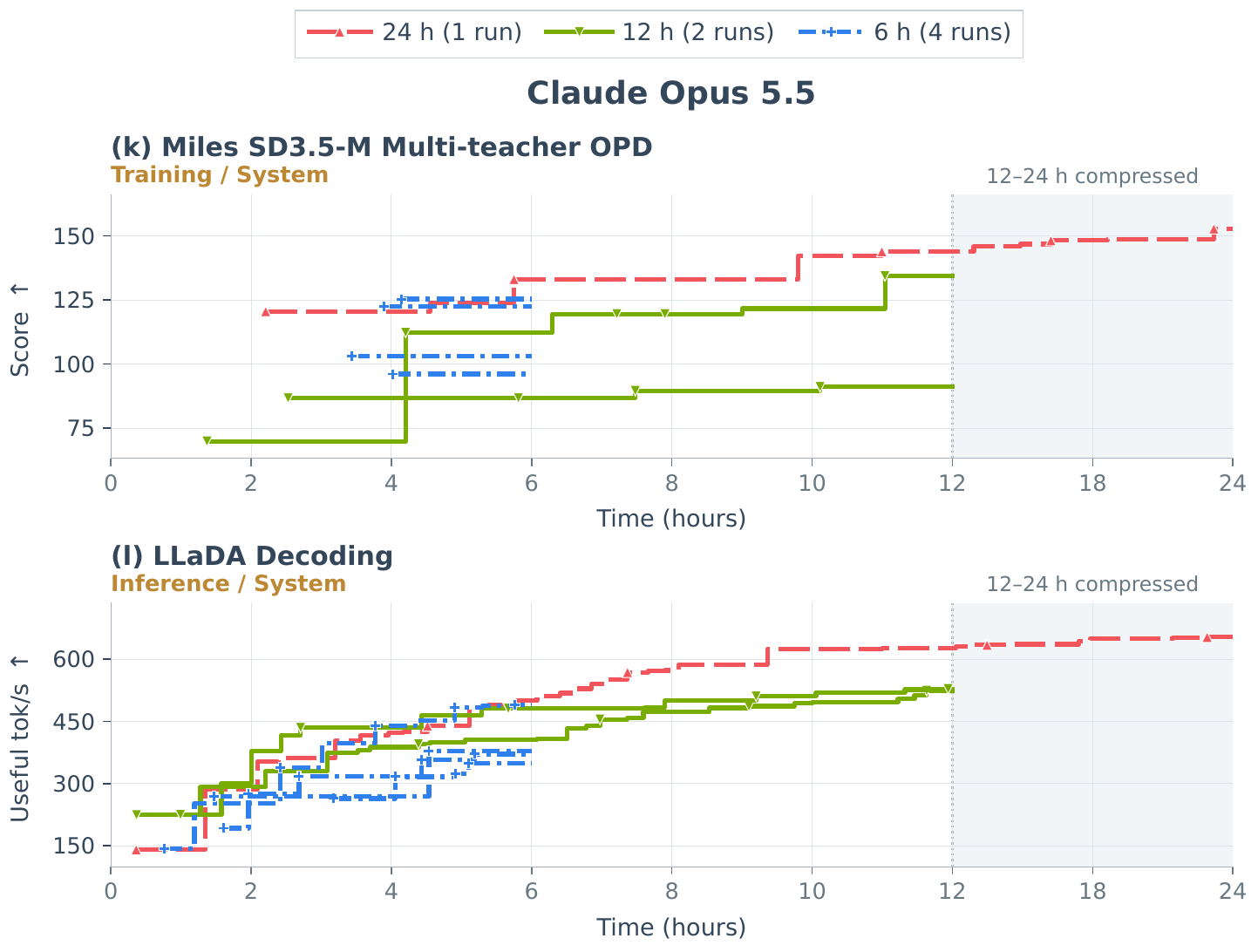}
    \caption{\textbf{Individual system trajectories with Claude Opus 5.5.} Miles SD3.5-M Multi-teacher OPD and LLaDA Decoding each use one 24-hour run, two 12-hour runs, and four 6-hour runs. Curves show performance over time. The 12--24-hour interval is compressed threefold; ticks show actual hours.}
    \label{fig:time_budget_system_claude}
\end{figure}

Figures~\ref{fig:time_budget_operator}--\ref{fig:time_budget_system_claude} show all 84 runs. Each panel shows one full-budget run, two half-budget runs, and four quarter-budget runs, aligned by elapsed time from each run's start. Runs with the same budget share their color, line style, and marker. Panel letters match Figure~\ref{fig:long_vs_short_runs}.

Claude achieves higher scores with short-horizon TTS on both operator tasks and Qwen3 ORPO, whereas Astra performs better with long-horizon optimization on all six tasks. In four task--model comparisons, short-horizon TTS leads at the end of its budget but is later overtaken by long-horizon optimization. On KVzip KV Scoring, both models trail their best two-run result at 12 hours, then surpass it at 13.90 hours for Astra and 12.67 hours for Claude.

\section{Benchmark Coverage and Horizon Comparison}
\label{app:benchmark_horizons}

This appendix documents the coverage and execution settings underlying Table~\ref{tab:benchmark_horizons}. A check denotes coverage or support documented in the inspected paper or repository; a cross denotes no documented coverage. Infra only requires an exclusive focus on AI infrastructure. The GPU tasks requires at least one task in the open-source release that explicitly uses a GPU. Long horizon requires the framework to support at least 12 hours of continuous execution on every task, with mechanisms that prevent voluntary early completion and recover interrupted sessions within a run.

We use the definitions in Appendix~\ref{app:optimization_levels} to assign task levels in Table~\ref{tab:benchmark_horizons}. A check requires at least one independently specified task at that level. Components used inside a higher-level task do not count as separate tasks. The same definitions apply to non-AI tasks in general-purpose benchmarks.

\subsection{Benchmarks in the Main Comparison}
\label{app:horizon_table_benchmarks}

\paragraph{KernelBench~\citep{Ouyang2025KernelBenchCL}.}
KernelBench evaluates generated GPU implementations of PyTorch computations using correctness and speedup over PyTorch Eager. Its individual-operation and simple fusion tasks establish operator coverage. Its Level 3 tasks include independent Inception, ResNet, and Transformer blocks, with interacting branches, attention, normalization, and feed-forward components; the benchmark permits the agent to choose the kernel granularity. These establish module coverage. Level 4 tasks evaluate Hugging Face model forward calls that return logits. Its refinement study uses 10 model calls with code and execution feedback; no numerical wall-clock cap is reported.

\paragraph{KernelBenchX~\citep{Wang2026KernelBenchXAC}.}
KernelBenchX evaluates correctness, speedup, and hardware efficiency across operators and GPUs. Its tasks specify individual GPU kernels, including fused computations. Its KernelAgent setting uses three workers with at most five refinement rounds each; its GEAK setting uses three iterations with four candidate solutions per round. The core KernelAgent implementation stops when any worker passes its tests or all workers exhaust their rounds. Test and seed generation also use model calls.

\paragraph{FlashInfer-Bench~\citep{Xing2026FlashInferBenchBT}.}
FlashInfer-Bench evaluates inference kernels on workloads derived from serving traces. The optimization targets are individual GPU kernels, including fused MoE kernels. The sequential setting uses 10 model calls with execution feedback and returns the best solution that passes its tests. No numerical wall-clock cap is reported.

\paragraph{RE-Bench~\citep{Wijk2024REBenchEF}.}
RE-Bench compares agents and human experts on seven research-engineering tasks, with up to 8 hours per attempt. Its scope includes ML research beyond infrastructure. Its prefix-sum GPU kernel establishes operator coverage, and optimization of the complete LLM Foundry finetuning script establishes system coverage. The restricted-architecture task allows designing and training an entire masked language model, while Fix Embedding scores a repaired model checkpoint and permits further training. Runs end when the agent finishes or reaches the time limit. The paper excludes pauses for API issues from agents' time usage.

\paragraph{InferenceBench~\citep{Yeon2026InferenceBenchAB}.}
InferenceBench requires a running OpenAI-compatible inference service on an H100 GPU. All four scenarios evaluate complete services. It uses 2-hour main budgets and a 1/2/4/8-hour time ablation. Its runners resume sessions using the remaining time, with a 300-second grace period added to the configured budget. Errors or exhausted retries can end a run early.

\paragraph{$\Phi$-Bench~\citep{Ding2026PhiBenchCL}.}
The benchmark covers kernel completion, long-horizon implementation, and end-to-end optimization. An audit of all 85 task manifests finds 43 tasks declaring GPU resources. Kernel implementations establish operator coverage. The priority-request-queue task permits changes only to the queue implementation, with exact ordering behavior and the surrounding vLLM code fixed; the memory-accounting task restricts edits to its accountant and scheduler. These establish module coverage. Complete nanoGPT training and vLLM serving tasks establish system coverage. The 85 tasks use caps of 1.5, 4, 5, or 6 hours. All 20 LHI (long-horizon implementation) tasks belong to the 1.5-hour group. The runner applies a timeout to a single agent invocation and ends the attempt when it returns.

\paragraph{EdgeBench~\citep{Zhu2026EdgeBenchUS}.}
EdgeBench covers science, ML, software, systems, math optimization, knowledge work, mathematics, and games. The paper's task descriptions establish operator coverage through a VLIW/SIMD kernel on a custom architecture simulator, module coverage through the EnTT graph component, and system coverage through a PostgreSQL-compatible server. These include non-AI tasks. The released official experiment configuration and container backends allocate CPU and memory resources without GPU device requests; several ML tasks also explicitly require CPU-only execution. Its main runs use 12 hours; extended experiments reach 28 and 72 hours. The released harness blocks voluntary stops and resumes eligible exits using the remaining execution budget, establishing framework support for long-horizon execution. Runs may end at the time or resume limit, or after exits that do not qualify for recovery.

\section{Quant-VideoGen KV Codec Case Study Details}
\label{app:qvg_case_details}

This appendix gives the experimental settings, codec designs, optimization steps, and metric definitions for Section~\ref{sec:qvg_case_study}.

\subsection{Experimental Setup}
\label{app:qvg_setup}

We compare the human codec with Claude Opus 5.5's codec on two video models, using 200 samples per dataset. Both codecs and the uncompressed BF16 reference use the same inputs and generation settings (Table~\ref{tab:qvg_setup}).

\begin{table}[!htbp]
    \centering
    \small
    \setlength{\tabcolsep}{5pt}
    \renewcommand{\arraystretch}{1.15}
    \caption{\textbf{Video generation and evaluation settings.} Frame indices start at zero.}
    \label{tab:qvg_setup}
    \begin{tabularx}{\textwidth}{@{}l>{\raggedright\arraybackslash}X>{\raggedright\arraybackslash}X@{}}
    \toprule
    \textbf{Setting} & \textbf{HY-WorldPlay} & \textbf{LongCat-Video} \\
    \midrule
    Dataset & WorldScore~\citep{Duan2025WorldScoreAU} & MovieGen~\citep{Polyak2024MovieGA} \\
    Input & Image, prompt, and camera actions & Prompt and fixed 73-frame prefix \\
    Resolution; seed & $832\times480$; 0 & $832\times480$; 0 \\
    Output frames & 285 & 273: 73 fixed and ten 20-frame segments \\
    Frame rate & 24 fps & 15 fps \\
    Denoising & Four steps & 50 steps; guidance scale 4.0 \\
    PSNR/SSIM/LPIPS frames & 29--44 & 73--92 \\
    \bottomrule
    \end{tabularx}
\end{table}

\textbf{WorldScore selection.} We map camera actions from the static WorldScore dataset to HY-WorldPlay controls and extend generation to 285 frames. Screening for severe collapse in uncompressed videos retains 200 of 289 reviewed samples. The screen uses only uncompressed references, except for eight pilot samples whose codec results had already been inspected; two of these were excluded. The retained set contains 131 outdoor and 69 indoor samples, and 115 photorealistic and 85 stylized samples. Its prompts do not overlap with the 11 original HY-WorldPlay task prompts. HY-WorldPlay uses our 8B integration with the official quantization code.

\textbf{MovieGen selection.} We sample 200 prompts from the 1,003-entry list released with Self-Forcing, excluding the six private-test and two public-test prompts, without screening generation quality.

\subsection{Codec Designs}
\label{app:qvg_codecs}

Table~\ref{tab:qvg_codec_comparison} compares the two codecs. For one attention head, $X\in\mathbb{R}^{S\times D}$ denotes either the cached keys (K) or values (V), with $S$ tokens and head dimension $D$. Let $x_i$ be token $i$'s vector, $\widehat{x}_i$ its reconstruction, $C$ the number of clusters, and $a_i$ its assigned cluster.

\begin{table}[!htbp]
    \centering
    \small
    \setlength{\tabcolsep}{5pt}
    \renewcommand{\arraystretch}{1.15}
    \caption{\textbf{Codec designs for Human and Claude Opus 5.5.}}
    \label{tab:qvg_codec_comparison}
    \begin{tabularx}{\textwidth}{@{}l>{\raggedright\arraybackslash}X>{\raggedright\arraybackslash}X@{}}
    \toprule
    \textbf{Component} & \textbf{Human} & \textbf{Claude Opus 5.5} \\
    \midrule
    K/V clustering & Separate assignments & Shared assignments; separate when $2D>256$ \\
    Cluster count & 256 & Adaptive (Equation~\ref{eq:qvg_joint_clustering}) \\
    Lloyd iterations & Two for HY-WorldPlay; up to 100 for LongCat-Video & Ten \\
    Residual transform & None & PCA, retaining all dimensions \\
    Quantization & Ternary values; one FP8 scale per 64 elements & Eight-state trellis quantization \\
    Residual storage & Two bits per symbol & rANS entropy coding \\
    \bottomrule
    \end{tabularx}
\end{table}

\textbf{Human codec.} Given cluster centers $c_j$, the codec~\citep{Xi2026QuantVA} assigns each token to its nearest center and quantizes the residual:
\begin{equation}
    a_i=\arg\min_j\|x_i-c_j\|_2^2,
    \qquad
    \widehat{x}_i=c_{a_i}+Q_3(x_i-c_{a_i}).
    \label{eq:qvg_human_codec}
\end{equation}
Here $Q_3$ rounds each residual value to $\{-s,0,s\}$, with FP8 scale $s$.

\textbf{Claude's codec.} Let $\bar{k}_i$ and $\bar{v}_i$ be the mean-centered, root-mean-square-normalized K and V vectors, and $d_j$ a joint cluster center. The cluster count and shared assignment are
\begin{equation}
    \begin{aligned}
        C &= \max(1,\min(256,\lfloor S/128\rfloor)), \\
        u_i &= [1.3\bar{k}_i,\bar{v}_i],
        \qquad a_i=\arg\min_j\|u_i-d_j\|_2^2.
    \end{aligned}
    \label{eq:qvg_joint_clustering}
\end{equation}
For each of K and V, let $\mu$ be the mean token vector, $\widetilde{c}_j$ a stored quantized center, and $B\in\mathbb{R}^{D\times D}$ the PCA residual transform. With transformed residual $z_i$ and trellis quantizer $Q_{\mathrm{TCQ}}$ of step size $\Delta$, reconstruction is
\begin{equation}
    \begin{aligned}
        z_i &= (x_i-\mu-\widetilde{c}_{a_i})B^{-1}, \\
        \widehat{x}_i &= \mu+\widetilde{c}_{a_i}+Q_{\mathrm{TCQ}}(z_i;\Delta)B.
    \end{aligned}
    \label{eq:qvg_agent_codec}
\end{equation}
$B$ is stored as an INT8 basis with scales. Both $B$ and $\Delta$ include the scaling factors for K, V, and each head. The trellis minimizes squared error after clipping, without a rate penalty. Range asymmetric numeral systems (rANS) coding losslessly encodes the resulting indices using their empirical frequencies. Stored data include centers, transforms, scales, shared assignments, and frequency tables.

\textbf{Task scores.} The human codec has a quality-weighted task score of 6.82, compared with 28.45 for Claude's final submission.

\subsection{Optimization Steps}
\label{app:qvg_optimization}

Table~\ref{tab:qvg_optimization_steps} maps the five numbered points in Figure~\ref{fig:qvg_case_study} to Claude Opus 5.5's submissions and explains the changes leading to each point. Scores are recomputed using the quality-weighted metric in Equation~\ref{eq:qvg_task_score}.

\begin{table}[!htbp]
    \centering
    \small
    \setlength{\tabcolsep}{5pt}
    \renewcommand{\arraystretch}{1.15}
    \caption{\textbf{Claude's optimization steps on Quant-VideoGen KV Codec.} Point numbers match Figure~\ref{fig:qvg_case_study}.}
    \label{tab:qvg_optimization_steps}
    \begin{tabularx}{\textwidth}{@{}cc>{\raggedright\arraybackslash}Xr@{}}
    \toprule
    \textbf{Point} & \textbf{Submission} & \textbf{Optimization} & \textbf{Score} \\
    \midrule
    1 & 6th & \textbf{4-bit quantization.} Claude tests a simple 4-bit codec and obtains its first valid private test score. & $5.80$ \\
    \addlinespace[4pt]
    2 & 7th & \textbf{PCA + entropy coding.} Claude returns to clustering and PCA residual quantization with rANS entropy coding. Video tests in the Work environment guide changes to the quantization step and encoded symbol set. & $14.07$ \\
    \addlinespace[4pt]
    3 & 8th & \textbf{Quantization-step tuning.} Claude uses video tests in the Work environment to select larger quantization steps for K and V, increasing compression while passing private tests. & $17.66$ \\
    \addlinespace[4pt]
    4 & 16th & \textbf{Trellis quantization.} Claude jointly quantizes PCA coefficients and uses finer steps for high-variance key directions. & $21.90$ \\
    \addlinespace[4pt]
    5 & 22nd & \textbf{Metadata compression + per-head quantization.} Claude losslessly compresses metadata and shares K/V cluster assignments. It also refines clustering and uses finer quantization steps for attention heads with larger residual magnitudes. & $28.45$ \\
    \bottomrule
    \end{tabularx}
\end{table}

\subsection{Evaluation Metrics}
\label{app:qvg_metrics}

Table~\ref{tab:qvg_metrics} summarizes the four metrics reported in Section~\ref{sec:qvg_case_study}. The results tables report arithmetic means over samples. PSNR, SSIM, and LPIPS compare raw RGB frames with the corresponding uncompressed BF16 frames, using the windows in Table~\ref{tab:qvg_setup}; frame scores are averaged within each window.

\begin{table}[!htbp]
    \centering
    \small
    \setlength{\tabcolsep}{5pt}
    \renewcommand{\arraystretch}{1.14}
    \caption{\textbf{Metrics for the codec comparison.} Arrows indicate the preferred direction.}
    \label{tab:qvg_metrics}
    \begin{tabularx}{\textwidth}{@{}l>{\raggedright\arraybackslash}X>{\raggedright\arraybackslash}X@{}}
    \toprule
    \textbf{Metric} & \textbf{What it measures} & \textbf{Computation or model} \\
    \midrule
    Compression $\uparrow$ & KV storage savings & Ratio of original to stored bytes (Equation~\ref{eq:qvg_compression_metric}) \\
    PSNR $\uparrow$ & Pixel reconstruction accuracy & Mean squared error, expressed in dB (Equation~\ref{eq:qvg_psnr}) \\
    SSIM $\uparrow$ & Local luminance, contrast, and structural similarity & Uniform $11\times11$ windows on each RGB channel (Equation~\ref{eq:qvg_ssim}) \\
    LPIPS $\downarrow$ & Perceptual difference from the BF16 reference & Weighted VGG feature distances; LPIPS version 0.1 \\
    \bottomrule
    \end{tabularx}
\end{table}

\textbf{Compression ratio.} For encoding call $j$, let $b_j$ be the original BF16 KV bytes and $\widehat{b}_j$ the stored bytes, including all reconstruction metadata. A video's compression ratio is
\begin{equation}
    R=\frac{\sum_j b_j}{\sum_j\widehat{b}_j}.
    \label{eq:qvg_compression_metric}
\end{equation}
Shared storage is counted once within each call. Only KV tensors passed to the codec enter this ratio. The task score combines compression and video quality for each video model (Appendix~\ref{app:task_qvg}).

\textbf{PSNR.} Let $I_t,\widehat{I}_t\in[0,1]^{H\times W\times3}$ be reference and reconstructed RGB frames at time $t$, with height $H$ and width $W$. Peak signal-to-noise ratio is
\begin{equation}
    \mathrm{MSE}_t=\frac{\|I_t-\widehat{I}_t\|_F^2}{3HW},
    \qquad
    \mathrm{PSNR}_t=-10\log_{10}\!\left(\max(\mathrm{MSE}_t,10^{-10})\right).
    \label{eq:qvg_psnr}
\end{equation}

\textbf{SSIM.} Structural similarity~\citep{Wang2004ImageQA} compares corresponding local windows $x$ and $y$. Let $\mu_x,\mu_y$ be their means, $\sigma_x^2,\sigma_y^2$ their variances, and $\sigma_{xy}$ their covariance. The local score is
\begin{equation}
    \mathrm{SSIM}(x,y)=
    \frac{(2\mu_x\mu_y+C_1)(2\sigma_{xy}+C_2)}
    {(\mu_x^2+\mu_y^2+C_1)(\sigma_x^2+\sigma_y^2+C_2)},
    \label{eq:qvg_ssim}
\end{equation}
with $C_1=0.01^2$ and $C_2=0.03^2$. We average the local scores over all pixels and RGB channels, using $11\times11$ uniform windows with stride 1 and zero padding of 5 pixels.

\textbf{LPIPS.} Learned perceptual image patch similarity~\citep{Zhang2018TheUE} measures weighted distances between VGG features. We use version 0.1 and rescale RGB inputs from $[0,1]$ to $[-1,1]$.

\paragraph{Quality-weighted task score.}
\phantomsection
\label{app:qvg_quality_score}
We rescore submitted solutions using compression and video quality. For each video model $m$, $R_m$ is the ratio of total BF16 to stored cache bytes, and quality metrics are averaged over test cases.

For frame similarity, we measure PSNR, SSIM, and LPIPS on the first segment generated with compressed KV. Let $\mathbf{x}_m$ contain their mean values and $\mathbf{x}^{\mathrm{Human}}_m$ the corresponding values for the human codec. We normalize each metric toward $\mathbf{b}=(40,1,0)$ and cap it at 1:
\begin{equation}
    f_m=\frac{1}{3}\sum_{j=1}^{3}
    \min\!\left(\frac{x_{m,j}-x^{\mathrm{Human}}_{m,j}}
    {b_j-x^{\mathrm{Human}}_{m,j}},\,1\right).
    \label{eq:qvg_fidelity_score}
\end{equation}
VBench~\citep{Huang2023VBenchCB} measures full-video quality. For dimension $k$, let $V_{m,k}$ and $V^{\mathrm{BF16}}_{m,k}$ be the mean scores with and without compression, and $\Delta_{m,k}$ the allowed decrease. We use the smallest normalized margin across its four dimensions:
\begin{equation}
    g_{m,k}=1-\frac{\max(V^{\mathrm{BF16}}_{m,k}-V_{m,k},\,0)}{\Delta_{m,k}},
    \qquad
    v_m=\min_k g_{m,k}-G_m^{\mathrm{Human}},
    \label{eq:qvg_vbench_score}
\end{equation}
where $G_m^{\mathrm{Human}}$ is the corresponding minimum margin for the human codec. We set $q_m=(f_m+v_m)/2$ in Equation~\ref{eq:qvg_task_score}.

\begin{table}[!htbp]
    \centering
    \small
    \setlength{\tabcolsep}{6pt}
    \renewcommand{\arraystretch}{1.12}
    \caption{\textbf{Human codec values and quality thresholds.} VBench tuples follow aesthetic quality, background consistency, imaging quality, and subject consistency.}
    \label{tab:qvg_score_reference}
    \begin{tabular}{@{}lcc@{}}
    \toprule
    \textbf{Quantity} & \textbf{LongCat-Video} & \textbf{HY-WorldPlay} \\
    \midrule
    $\mathrm{PSNR}^{\mathrm{Human}}$ (dB) & 23.6403 & 31.0633 \\
    $\mathrm{SSIM}^{\mathrm{Human}}$ & 0.8704 & 0.9483 \\
    $\mathrm{LPIPS}^{\mathrm{Human}}$ & 0.0964 & 0.0435 \\
    $G^{\mathrm{Human}}$ & 0.4060 & 0.3939 \\
    $\Delta$ & (1.873, 2.685, 6.174, 4.196) & (2.055, 1.511, 2.415, 1.984) \\
    \bottomrule
    \end{tabular}
\end{table}

\subsection{Additional Results}
\label{app:qvg_additional_results}

Table~\ref{tab:qvg_case_demo} gives all reported metrics; Table~\ref{tab:qvg_improvements_timing} reports how often Claude improves quality and separates encoding from decoding time.

\begin{table}[!htbp]
    \centering
    \footnotesize
    \setlength{\tabcolsep}{2.5pt}
    \renewcommand{\arraystretch}{1.12}
    \caption{\textbf{Human and Claude on WorldScore and MovieGen.} Entries are dataset means. Time is in seconds; MovieGen timing uses 188 matched pairs. VBench~\citep{Huang2023VBenchCB} columns report aesthetic quality, background consistency, imaging quality, and subject consistency.}
    \label{tab:qvg_case_demo}
    \begin{tabular*}{\textwidth}{@{\extracolsep{\fill}}lllrrrrrrrrr@{}}
    \toprule
    & & & \multicolumn{2}{c}{Efficiency} & \multicolumn{3}{c}{Frame similarity} & \multicolumn{4}{c}{Full-video VBench $\uparrow$} \\
    \cmidrule(lr){4-5}\cmidrule(lr){6-8}\cmidrule(l){9-12}
    Video model & Dataset & Codec & Comp. $\uparrow$ & Time $\downarrow$ & PSNR $\uparrow$ & SSIM $\uparrow$ & LPIPS $\downarrow$ & Aes. & Bg. & Img. & Subj. \\
    \midrule
    \multirow{2}{*}{HY-WorldPlay} & \multirow{2}{*}{WorldScore} & Human & 6.75 & 305 & 31.15 & 0.923 & 0.059 & 59.8 & 93.2 & 69.9 & 88.2 \\
     & & Claude & 20.89 & 262 & 36.33 & 0.975 & 0.023 & 61.8 & 94.0 & 70.9 & 89.6 \\
    \addlinespace[4pt]
    \multirow{2}{*}{LongCat-Video} & \multirow{2}{*}{MovieGen} & Human & 6.88 & 1734 & 23.67 & 0.836 & 0.126 & 63.8 & 93.0 & 65.8 & 89.1 \\
     & & Claude & 18.83 & 1760 & 26.95 & 0.884 & 0.085 & 66.0 & 94.2 & 68.4 & 91.2 \\
    \bottomrule
    \end{tabular*}
\end{table}

\begin{table}[!htbp]
    \centering
    \small
    \setlength{\tabcolsep}{6pt}
    \renewcommand{\arraystretch}{1.12}
    \caption{\textbf{Per-sample improvements and codec timing.} Counts compare Claude with the human codec. Times show Human $\rightarrow$ Claude in seconds per video.}
    \label{tab:qvg_improvements_timing}
    \begin{tabular*}{\textwidth}{@{\extracolsep{\fill}}lcc@{}}
    \toprule
    \textbf{Measurement} & \textbf{WorldScore} & \textbf{MovieGen} \\
    \midrule
    Higher compression & 200/200 & 200/200 \\
    Higher PSNR & 200/200 & 170/200 \\
    Higher SSIM & 200/200 & 169/200 \\
    Lower LPIPS & 200/200 & 176/200 \\
    Higher VBench scores (four dimensions) & 615/800 & 659/800 \\
    \midrule
    Mean total encoding time (s) & $67.3\rightarrow41.4$ & $103.2\rightarrow49.5$ \\
    Mean total decoding time (s) & $25.2\rightarrow8.6$ & $18.5\rightarrow102.4$ \\
    Decodes per compressed cache & 4 & $\approx56$ \\
    \bottomrule
    \end{tabular*}
\end{table}

\end{document}